\documentclass[10pt,twocolumn,letterpaper]{article}

\usepackage[pagenumbers]{cvpr}    

\usepackage{graphicx}
\usepackage{amsmath}
\usepackage{amssymb}
\usepackage{booktabs}
\usepackage[utf8]{inputenc} 
\usepackage[T1]{fontenc}    
\usepackage{hyperref}       
\usepackage{url}            
\usepackage{booktabs}       
\usepackage{amsfonts}       
\usepackage{nicefrac}       
\usepackage{microtype}      
\usepackage{xcolor}         
\usepackage{hhline}
\usepackage{booktabs}
\usepackage{algorithm}
\usepackage[noend]{algpseudocode}
\usepackage{subcaption}
\usepackage{pifont}
\usepackage{amsfonts}
\usepackage{amsmath}
\usepackage{arydshln}
\usepackage{multirow}
\usepackage{multicol}
\usepackage[export]{adjustbox}
\usepackage[normalem]{ulem}
\usepackage{xspace}
\usepackage{rotating}
\usepackage{xhfill}
\usepackage{soul}

\usepackage[shortlabels]{enumitem}

\newcommand{\nl}[1]{\textit{``#1''}}

\usepackage[capitalize]{cleveref}
\Crefname{section}{Section}{Sections}
\crefname{section}{Sec.}{Secs.}
\Crefname{table}{Table}{Tables}
\crefname{table}{Tab.}{Tabs.}
\crefname{figure}{Fig.}{Figs.}
\crefname{figure}{Fig.}{Figs.}
\crefname{algocf}{alg.}{algs.}
\Crefname{algocf}{Alg.}{Algs.}

\makeatletter
\DeclareRobustCommand\onedot{\futurelet\@let@token\@onedot}
\def\@onedot{\ifx\@let@token.\else.\null\fi\xspace}

\def\eg{\emph{e.g}\onedot}

\def\ie{\emph{i.e}\onedot}

\def\etc{\emph{etc}\onedot}

\definecolor{mydarkgreen}{rgb}{0.02,0.6,0.02}
\definecolor{jazzberryjam}{rgb}{0.65, 0.04, 0.37}

\usepackage[capitalize]{cleveref}
\crefname{section}{Sec.}{Secs.}
\Crefname{section}{Section}{Sections}
\Crefname{table}{Table}{Tables}
\crefname{table}{Tab.}{Tabs.}

\def\cvprPaperID{*****} 
\def\confName{CVPR}
\def\confYear{2022}

\begin{document}

\title{The Hidden Language of Diffusion Models}
\author{\vspace{0.2cm}Hila Chefer$^*$\textsuperscript{\rm 1,2} \hspace{0.65cm} Oran Lang$^1$ \hspace{0.65cm} Mor Geva$^3$ \hspace{0.65cm} Volodymyr Polosukhin$^1$  \hspace{0.65cm} Assaf Shocher$^1$ \vspace{-0.2cm}\\
\hspace{0.65cm}Michal Irani$^{1,4}$ \hspace{0.65cm} Inbar Mosseri$^1$ \hspace{0.65cm} Lior Wolf$^2$\vspace{0.25cm}  \\
\textsuperscript{\rm 1}Google Research \quad \textsuperscript{\rm 2}Tel-Aviv University \quad \textsuperscript{\rm 3}Google DeepMind \quad \textsuperscript{\rm 4}Weizmann Institute \vspace{0.2cm} \\
\small\url{https://hila-chefer.github.io/Conceptor/} \vspace{-0.8cm}
}

\twocolumn[{%
\vspace{-1em}
\maketitle
\renewcommand\twocolumn[1][]{#1}%
\vspace{-0.1in}
\begin{center}
    \centering
    \includegraphics[width=\linewidth]{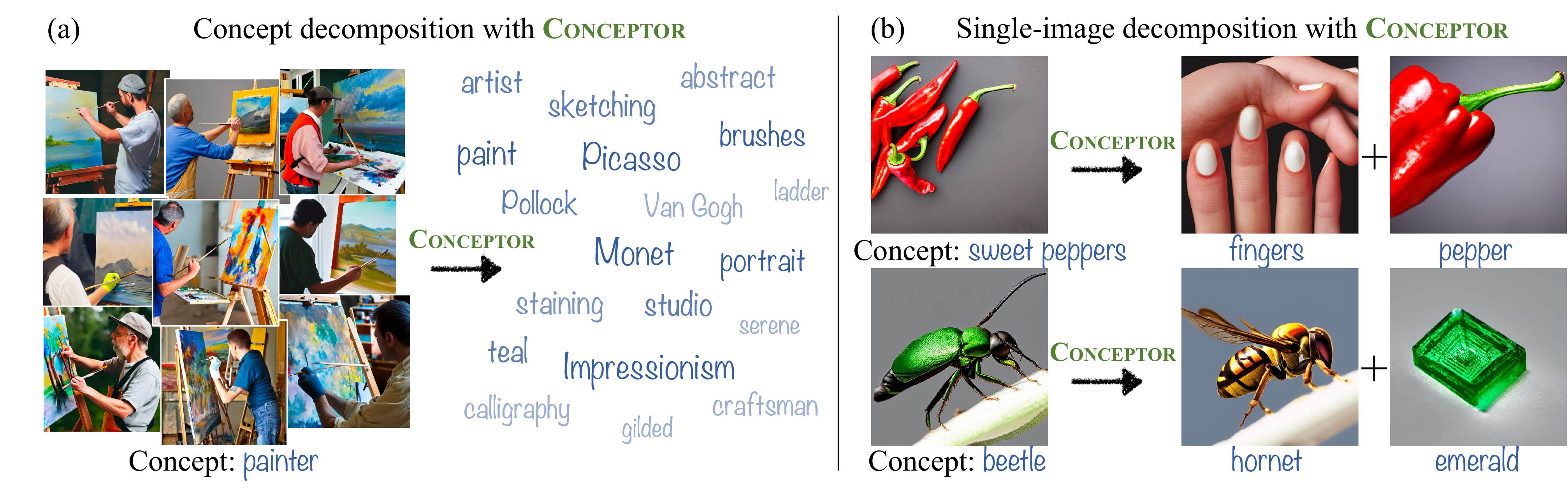}
    \vspace{-0.5cm}
    \captionof{figure}{
    Concept interpretation with \textsc{Conceptor}. (a) Given a set of representative concept images, \textsc{Conceptor} learns to decompose the concept into a weighted combination of interpretable elements (font sizes indicate weights). The decomposition exposes interesting behaviors such as reliance on prominent painters and renowned artistic styles (\eg, \nl{Monet}, \nl{Impressionism}). (b) Given a \emph{specific} generated image, \textsc{Conceptor} extracts its primary contributing elements, revealing surprising visual connections (\eg, \nl{sweet peppers} are linked to \nl{fingers} due to their common shape). 
    }
    \label{fig:teaser}
\end{center}%
}]

\maketitle

\def\thefootnote{*}\footnotetext{The first author performed this work as an intern at Google Research.}

\begin{abstract}

Text-to-image diffusion models have demonstrated an unparalleled ability to generate high-quality, diverse images from a textual prompt. However, the internal representations learned by these models remain an enigma. In this work, we present  \textsc{Conceptor}, a novel method to interpret the internal representation of a textual concept by a diffusion model. This interpretation is obtained by decomposing the concept into a small set of human-interpretable textual elements. Applied over the state-of-the-art Stable Diffusion model, \textsc{Conceptor} reveals non-trivial structures in the representations of concepts. For example, we find surprising visual connections between concepts, that transcend their textual semantics. We additionally discover concepts that rely on mixtures of exemplars, biases, renowned artistic styles, or a simultaneous fusion of multiple meanings of the concept.
Through a large battery of experiments, we demonstrate \textsc{Conceptor}'s ability to provide meaningful, robust, and faithful decompositions for a wide variety of abstract, concrete, and complex textual concepts, while allowing to naturally connect each decomposition element to its corresponding visual impact on the generated images.
\end{abstract}

\section{Introduction}
Generative models have demonstrated unprecedented capabilities to create high-quality, diverse imagery based on textual descriptions~\cite{Balaji2022eDiffITD,gafni2022make,ramesh2021zero,rombach2022high,Saharia2022PhotorealisticTD}. While revolutionary, recent works have demonstrated that these models often suffer from heavy reliance on biases~\cite{chuang2023debiasing, luccioni2023stable} and occasionally also data memorization~\cite{Carlini2023ExtractingTD, somepalli2022diffusion}. However, all these works draw conclusions by a simple external evaluation of the output images, while research on understanding the internal representations learned by the model remains scarce. Thus, our understanding of these impressive models remains limited.

In this work, we aim to develop a method to interpret the inner representations of text-to-image diffusion models. Our approach draws inspiration from the field of concept-based interpretability~\cite{li-etal-2021-implicit, Kim2017InterpretabilityBF,Ghorbani2019TowardsAC}, which proposes to interpret the model's decision-making for a given input by \emph{decomposing} it into a set of elements that impact the prediction (\eg, decomposing \nl{cat} into \nl{whiskers}, \nl{paws}, \etc). Under this setting, an ``interpretation'' can be considered to be a \emph{mapping function} from the internal state of the model to a set of concepts humans can understand~\cite{Kim2017InterpretabilityBF}.
Notably, existing approaches for concept-based interpretability are not directly applicable to generative models, and, as we demonstrate, fall short of producing meaningful interpretations for such models. Therefore, we propose a novel method to produce concept-based explanations for diffusion models, which leverages their unique structure and properties.

Our method, \textsc{Conceptor}, learns a \emph{pseudo-token}, realized as a combination of interpretable textual elements (\cref{fig:teaser}(a)). To obtain the pseudo-token, \textsc{Conceptor} trains a neural network to map each word embedding in the vocabulary of the model to a corresponding coefficient, with the objective of denoising the concept images. The pseudo-token is then constructed as a linear combination of the top vocabulary elements weighted by their learned coefficients. This formulation allows us to exploit both the model's powerful ability to link between text and image, and the rich semantic information encapsulated in the text encoder. Through a large battery of experiments, we demonstrate that \textsc{Conceptor} produces decompositions that are meaningful, robust, and faithful to the model.

We use \textsc{Conceptor} to analyze the state-of-the-art text-to-image diffusion model, Stable Diffusion~\cite{rombach2022high} across various concepts, including concrete (\eg, \nl{a nurse}), abstract (\eg, \nl{affection}) and complex (\eg, \nl{elegance on a plate}) concepts, as well as homograph concepts (\eg, \nl{a crane}). 
\textsc{Conceptor} reveals many interesting observations about the learned representations. (i) As demonstrated in~\cref{fig:teaser}(b), \textsc{Conceptor} can be used to decompose a generated image to its own subset of driving elements. We find non-trivial combinations of features that control different visual aspects such as shapes, textures, and colors. (ii) We observe that some concepts such as \nl{a president} or \nl{a rapper} are represented mostly by \emph{exemplars}, \ie, well-known instances from the concept, such that the generated images are \emph{interpolations} of those instances.
(iii) We additionally find that consistent with previous work~\cite{rassin2022dalle}, the model learns to mix the multiple meanings of homograph concepts. We expand those findings and discover cases where these meanings are leveraged \emph{simultaneously}, creating images that mix both meanings in a \emph{single object}. 
(iv) Finally, we demonstrate our method's effectiveness in the detection of non-trivial biases. 

To conclude, our work makes the following contributions:
 \begin{itemize} [leftmargin=*,topsep=0pt,itemsep=0pt,parsep=0pt]
    \item We present \textsc{Conceptor}, a novel method to decompose a textual concept into a set of interpretable elements. Our method utilizes a unique form of decomposition in which a linear combination is learned as a mapping from the textual embedding space to a coefficient.

    \item We demonstrate profound learned connections between concepts that transcend textual correlations. 

    \item We discover non-trivial structures in the learned decompositions such as interpolations of exemplars, reliance on renowned artistic styles, and mixing of different meanings of the concept.

    \item We demonstrate the detection of biases that are not easily observable visually. These observations can help discuss important ethical questions on a factual basis.
 \end{itemize}
\section{Related Work}
\label{sec:related}
\noindent{\bf Text-guided image generation\quad}
Recently, impressive results were achieved for text-guided image generation with large-scale auto-regressive models~\cite{ramesh2021zero,yu2022scaling} and diffusion models~\cite{Ramesh2022Hierarchical,nichol2021glide,rombach2022high,Saharia2022PhotorealisticTD}.
In the context of text-to-image diffusion models, a related line of work aims to introduce personalized concepts to a pre-trained text-to-image model by learning to map a set of images to a ``token'' in the text space of the model~\cite{gal2022image, ruiz2022dreambooth,kumari2022multi}. Importantly, these methods do not produce decomposable or interpretable information and mostly result in a rigid learned vector that resides outside of the model's distribution~\cite{Voyov2023Pplus}.

\noindent{\bf Concept-based interpretability\quad}
A similar analysis to ours was conducted on concept representations in the context of language models~\cite{patel2022mapping, li-etal-2021-implicit, li2023emergent, lovering-pavlick-2022-unit}, often through projections to the vocabulary space~\cite{geva-etal-2022-transformer, geva2023dissecting, ram2022you}. For image classifiers based on CNNs, TCAV~\cite{Kim2017InterpretabilityBF} propose the first concept-based explainability method by training a linear classifier over user-defined concepts. ACE~\cite{Ghorbani2019TowardsAC} leverages multi-resolution segmentation of the class images and clusters the bottleneck representations of the crops. ICE~\cite{Zhang2020InvertibleCE} and CRAFT~\cite{Fel2022CRAFTCR} apply matrix factorization methods on a feature map matrix extracted from a set of patches to obtain the decomposition. 
Note that all these methods are not applicable directly to diffusion models. First, most of the methods are based on CNN architectures~\cite{Ghorbani2019TowardsAC} with non-negative activations~\cite{Zhang2020InvertibleCE,Fel2022CRAFTCR}, thus cannot be directly generalized to negative feature matrices. Most importantly, all methods above perform concept importance estimation by measuring the shift in prediction or by employing saliency methods, both are not trivially applicable to a generative model. 

\noindent{\bf Diffusion model interpretability\quad}
Shared text-image representations such as CLIP~\cite{radford2021learning} have been analyzed in the past~\cite{Chefer_2021_CVPR,Chefer_2021_ICCV, yun2023visionlanguage}. However, none of these works has been generalized to generative models. As far as we can ascertain, the closest effort to explaining text-to-image models is a simple visualization of the cross-attention maps~\cite{hertz2022prompt, han2023svdiff, chefer2023attend}.
Some works~\cite{rassin2022dalle, Carlini2023ExtractingTD} have attempted to investigate the images produced by text-to-image diffusion models, and have even found evidence of memorization~\cite{Carlini2023ExtractingTD}. However, these works rely entirely on the generated images, and are, therefore, time-consuming and require access to the data. For example~\cite{somepalli2022diffusion} were able to show memorization for less than $2\%$ of the generations with SD. Additionally, these works do not attempt to dissect the model's inner representations. 

\section{Method}
\label{sec:method}

\noindent{\bf Preliminaries: Latent Diffusion Models (LDMs)\quad}
\label{sec:background}
We apply our method over the state-of-the-art Stable Diffusion (SD) model~\cite{rombach2022high}. 
SD employs a denoising diffusion probabilistic model (DDPM)~\cite{sohl2015deep,ho2020denoising} over an input latent vector $z_T\sim\mathcal{N}(0,1)$ and gradually denoises it. 
Namely, at each timestep, $t=T,\dots, 1$, the DDPM receives a noised latent vector $z_t$ and produces a less noisy vector $z_{t-1}$, which serves as the input to the next step. 

During the denoising process, the model is typically conditioned on a text encoding for an input prompt $\mathcal{P}$, produced by a frozen CLIP text encoder~\cite{radford2021learning}, which we denote by $\mathcal{C}$. The text encoder converts the textual prompt $\mathcal{P}$ to a sequence of tokens, which can be words, sub-words, or punctuation marks. Then, the encoder's vocabulary, $\mathcal{V}\in \mathbb{R}^{N, d}$, is used to map each token in the prompt to an embedding vector $w\in \mathbb{R}^d$, where $d$ is the embedding dimension of the encoder, and $N$ is the number of tokens in the vocabulary. 
The DDPM model is trained to minimize the loss,
\begin{equation}
    \mathcal{L}_{rec} = \mathbb{E}_{z,\mathcal{P},\varepsilon\sim\mathcal{N}(0,1),t} \left [ || \varepsilon - \varepsilon_\theta(z_t, t, \mathcal{C}(\mathcal{P})) ||^2 \right ],
    \label{eq:reconstruction}
\end{equation}
for, 
\begin{equation}
    z_t = \sqrt{\alpha_t}z + \sqrt{1-\alpha_t}\varepsilon,
    \label{eq:latent}
\end{equation}
where $\varepsilon_\theta$ is a trained UNet~\cite{ronneberger2015u}, and $0=\alpha_T < \alpha_{T-1} <  \dots <\alpha_0=1$. In other words, during training, the input image $x$ is encoded to its corresponding latent vector $z$. A noise vector $\varepsilon$ and a timestep $t$ are drawn randomly. The noise vector $\varepsilon$ is then added to the latent vector $z$ as specified in Eq.~\ref{eq:latent}, and the UNet is trained to predict the added noise $\varepsilon$.

\noindent{\bf \textsc{Conceptor}\quad} 
Our goal is to interpret the internal representation of an input concept $c$ in a text-to-image diffusion model $\varepsilon_{\theta}$.
Formally, given a prompt $\mathcal{P}^c$ for the concept $c$, we learn a decomposition for the concept using the vocabulary $\mathcal{V}$. 
This decomposition is realized as a pseudo-token $w^* \notin \mathcal{V}$ that is constructed as a weighted combination of a subset of tokens from $\mathcal{V}$, \ie,
\begin{align}
    w^* &= \sum_{i=1}^n \alpha_i w_i  &  &\text{s.t.} &  w_i\in \mathcal{V}, \alpha_1, \dots, \alpha_n \geq 0
    \label{eq:pseudo}
\end{align}
where $n << N$ is a hyperparameter that determines the number of tokens to use in the combination. 

Learning the set of $n$ vocabulary elements $w_i$ and their associated coefficients $\alpha_i$ is done separately for each concept $c$. We begin by collecting a training set $\mathcal T$ of $100$ concept images. These images provide the statistics for the concept features we wish to learn. Next, we construct our learned pseudo-token over the vocabulary $\mathcal{V}$. This construction serves two purposes. First, note that the vocabulary $\mathcal{V}$ is \emph{expressive and diverse}, containing roughly $50,000$ tokens, ranging from famous personalities to emojis and even names of video games (\eg, \texttt{Fortnite} is a single token in $\mathcal{V}$). By learning our decomposition over this rich set of candidate concepts, we facilitate a meaningful yet non-restrictive optimization over semantic, human-understandable information. Thus, our choice of candidate concepts achieves the best of both worlds -- it is both expressive and interpretable. Second, we make a design choice to \emph{optimize the coefficients as a function of the word embeddings}. This choice is critical to the success of our method since it utilizes the rich textual embedding space of CLIP in determining the coefficients. Effectively, this reduces the optimization problem from optimizing $50,000$ unrelated coefficients to learning a mapping from a smaller semantic space, $\mathbb{R}^d$.
Specifically, our method assigns a coefficient $\alpha$ for each word embedding $w$ using a learned 2-layer MLP, which takes as input the word embedding vector $w$ as follows, 
\begin{align}
   \forall w\in \mathcal{V}: & \;\;\alpha = f(w) = \sigma\left(W_2(\sigma(W_1(w)))\right),
   \label{eq:mlp}
\end{align}
where $\sigma$ is the ReLU non-linearity~\cite{Agarap2018DeepLU}, and $W_1, W_2$ are linear mappings.
Based on $f$, we compute $w^*_N = \sum_{i=1}^N f(w_i) w_i $. Note that this pseudo-token is not identical to the output token $w^*$ since $w^*_N$ contains all tokens in $\mathcal{V}$. $w^*$ is obtained by the top tokens from $w^*_N$, as described in~Eq.~\ref{eq:sparsity}.

To learn a meaningful pseudo-token $w^*$, we optimize our MLP to reconstruct the images generated from $\mathcal{P}^c$. This choice encourages our pseudo-token to imitate the denoising process of the concept images.
We draw a random noise $\varepsilon\sim \mathcal{N}(0,1)$ and a timestep $t\in \{1,\dots, T\}$ for each image, and noise the images according to~Eq.~\ref{eq:latent}. We then employ the model's reconstruction objective from Eq.~\ref{eq:reconstruction}.

As mentioned, $w^*_N$ considers all the tokens in the vocabulary. 
However, for better interpretability, we wish to represent the input concept with a \textit{small} set of $n << N$ tokens. Notate by $w_1,\dots,w_n\in \mathcal{V}$ the tokens with the highest learned coefficients. We add a regularization loss to encourage the pseudo-token $w^*_N$ to be dominated by these top $n$ tokens, \ie,
\begin{equation}
    \mathcal{L}_{sparsity} = 1- cosine\left(w^*, w^*_N\right).
    \label{eq:sparsity}
\end{equation}
This encourages the pseudo-token $w^*$, defined by the top $n$ tokens in $\mathcal{V}$, to be semantically similar to $w^*_N$, which is defined by the entire vocabulary.
Our overall objective function is, therefore,
\begin{equation}
    \mathcal{L} = \mathcal{L}_{rec} + \lambda_{sparsity} \mathcal{L}_{sparsity},
    \label{eq:loss}
\end{equation}
In our experiments, we set $\lambda_{sparsity}=0.001, n=50$.
At inference time, we employ the MLP on the vocabulary $\mathcal{V}$ and consider the top $n=50$ tokens to compose $w^*$, as specified in~Eq.~\ref{eq:pseudo}. Implementation details and a figure describing our method can be found in~\cref{app:implementation}.

\begin{figure*}[t]
    \centering
    \addtolength{\belowcaptionskip}{-12pt}
    \includegraphics[width=1\linewidth, clip]{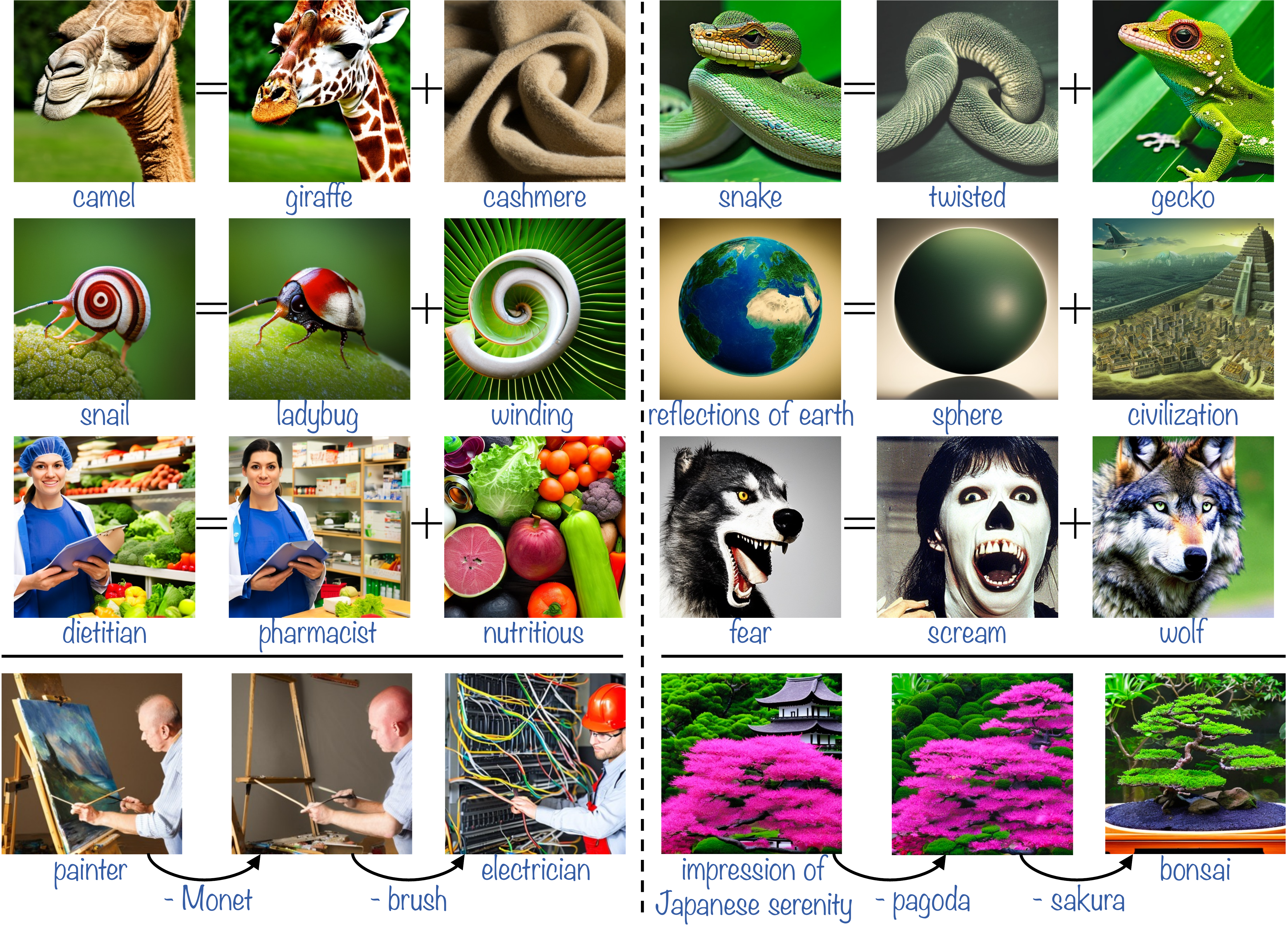}
    \vspace{-22pt}
    \caption{ 
    Decompositions of single images by \textsc{Conceptor}. The top rows present images found to contain two elements. The last row shows more complex mixtures by removing one element at a time. 
    The examples demonstrate the \emph{meaningfulness} of our learned decompositions.
    }
    \label{fig:single_image}
    \vspace{-2px}
\end{figure*}
\noindent{\bf Single-image decomposition\quad}
\label{sec:single_image}
Given an image $I$ that was generated by SD for a concept $c$, we wish to determine the subset of the tokens from the decomposition $w^*$, that drove the generation of this specific image. This is done via an iterative process over the tokens $w_j\in w^*$ as follows; at each step, we attempt to remove a single token from the decomposition, $w^*_j=\sum_{i\neq j} \alpha_i w_i$, and generate the corresponding image $I_j$ with the prompt $\mathcal{P}^{w^*_j}$ and the same seed. Next, we use CLIP's image encoder to determine if $I_j$ is semantically identical to $I$. If the CLIP score of the two images is higher than $95$, we remove $w_j$ from $w^*$ and continue to the next token. This criterion avoids incorporating tokens whose removal only causes minor non-semantic modifications to the image $I$ (such as a slight shift in pose). This process is repeated for all tokens in $w^*$ until no tokens are removed.

\begin{figure*}[t]
    \centering
    \includegraphics[width=1.01\linewidth, clip]{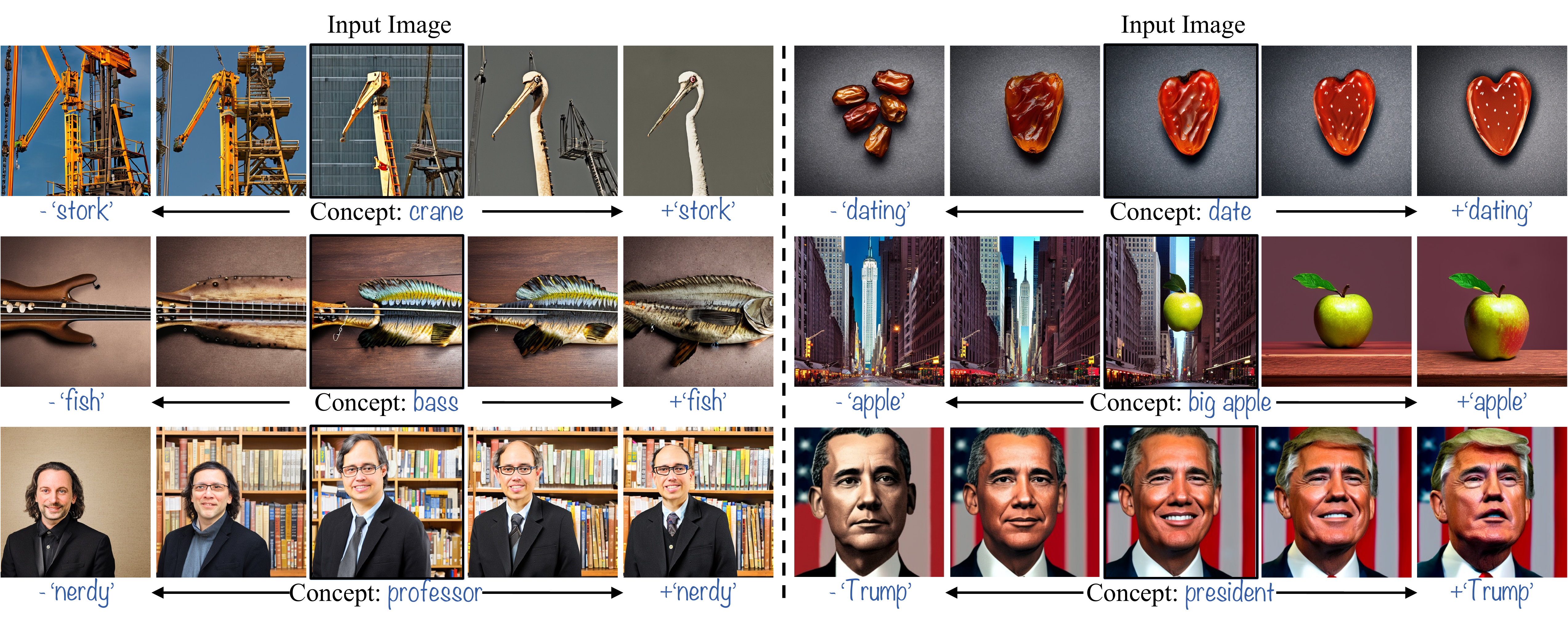}
    \vspace{-18px}
    \caption{ 
     Coefficient manipulation. For each of the listed concepts, we manipulate the coefficient of a single element from the decomposition and observe its visual impact on the generated image.
     The examples demonstrate the \emph{meaningfulness} of our learned decompositions.
    }
    \label{fig:manipulations}
    \vspace{-10px}
\end{figure*}
\section{Experiments}
\noindent{\bf Concept-based explanation desiderata\quad}
Following previous literature on concept-based explainability~\cite{Ghorbani2019TowardsAC}, we begin by defining a set of desired properties that concept-based explanations for diffusion models should uphold. These properties will be the criteria to evaluate our method. (i) \textbf{Meaningfulness}- Each decomposition element should be semantically meaningful and human-understandable. (ii) \textbf{Faithfulness}- The decomposition should be faithful to the concept representation by the model. In other words, the decomposition should reconstruct the features manifested in each of the concept images, and produce images that are in the distribution of the concept images. (iii) \textbf{Robustness}- The decomposition should be independent of the selection of the training set and the initialization.
Next, we conduct extensive experiments to demonstrate our method's ability to produce meaningful, faithful, and robust decompositions for diverse types of concepts. Throughout this section, we notate by $w^c$ the token(s) corresponding to the concept $c$.

\noindent{\bf Data\quad} We construct a diverse and comprehensive dataset of $188$ concepts, comprised of the basic classes from CIFAR-10~\cite{Krizhevsky2009LearningML}, a list of $28$ professions from the Bias in Bios dataset~\cite{1901.09451}, $10$ basic emotions and $10$ basic actions, all $30$ prompts from the website \emph{Best 30 Stable Diffusion Prompts for Great Images}\footnote{https://mspoweruser.com/best-stable-diffusion-prompts/}, which contains complex prompts that require hierarchical reasoning (\eg, \nl{Medieval village life}, \nl{impression of Japanese serenity}), and, finally, we consider $100$ random concepts from the ConceptNet~\cite{Speer2013ConceptNet5A} knowledge graph to allow for large-scale evaluation of the methods.
A full list of all concepts is provided in~\cref{app:data}.

\begin{table*}[t]
\small
\centering
\setlength{\tabcolsep}{1pt}
\addtolength{\belowcaptionskip}{-0pt}
\caption{Quantitative evaluation of \textsc{Conceptor} and the baselines.}
\vspace{-10px}
\begin{tabular}{l@{\hspace{0.1cm}} c@{\hspace{0.4cm}} c@{\hspace{0.4cm}} c@{\hspace{0.4cm}} c@{\hspace{0.4cm}} c} 
\toprule
 \multirow{1}{*}{{Method}}&  {CLIP pairwise$\uparrow$}& \multirow{1}{*}{\small{LPIPS}$\downarrow$}   & \multirow{1}{*}{FID per concept$\downarrow$} & \multirow{1}{*}{FID entire set$\downarrow$}  & \begin{tabular}{c} Token diversity \end{tabular}  \\
\midrule
\begin{tabular}{l}BLIP-2 token \end{tabular}  &  66.3 $\pm$ 16.8 & 0.60 $\pm$ 0.13 & 218.1 $\pm$ 93.4 &  46.6 & 52.1 $\pm$ 9.7  \\
\begin{tabular}{l} BLIP-2 sentence \end{tabular}  &  78.7 $\pm$ 11.2 & 0.57 $\pm$ 0.8 & 158.6 $\pm$ 74.0 & 23.7  & {65.8} $\pm$ 1.9 \\
\begin{tabular}{l} PEZ\end{tabular}  &  79.1 $\pm$ 9.9   &  0.56 $\pm$ 0.06 &  150.6 $\pm$ 63.4 &  18.2& \textbf{75.9} $\pm$ 1.2 \\
\begin{tabular}{l} NMF\end{tabular}  & 80.0 $\pm$ 9.7  & 0.53 $\pm$ 0.06 & 147.0 $\pm$ 68.9 & 21.6 &  ---  \\
\begin{tabular}{l} k-means\end{tabular}  &  82.5$\pm$ 8.4  & 0.53 $\pm$ 0.05& 132.5 $\pm$ 60.0 & 21.3 &  ---  \\
\begin{tabular}{l} PCA\end{tabular}  & 83.0 $\pm$ 8.0  & 0.53 $\pm$ 0.09 & 130.8 $\pm$ 52.9  & 19.8 &  ---  \\
\begin{tabular}{l} \textbf{\textsc{Conceptor}} \end{tabular}  & \textbf{86.2} $\pm$ 8.3 & \textbf{0.44} $\pm$ 0.09 & \textbf{109.5} $\pm$ 51.8  & \textbf{9.8} & {69.8} $\pm$ 3.4 \\
\bottomrule 
\end{tabular}
\label{tb:reconstruction}
\vspace{-12px}
\end{table*}
\subsection{Qualitative Results}
\subsubsection{From Text to Multi-Modal Explanations}
The ability to link between textual decomposition elements and their visual manifestation is important to facilitate human understanding and establish \emph{meaningfulness}. We propose two strategies to obtain these connections. First is our single-image decomposition scheme, described in~\cref{sec:single_image}. Second, we propose to gradually manipulate the element's coefficient to observe the visual changes it induces.

\noindent{\bf Single-image decomposition\quad}
\cref{fig:teaser,fig:single_image} and \cref{app:single_image} contain examples of decompositions over images generated by SD. The first three rows of \cref{fig:single_image} present examples of images that decompose into two elements,  
while the last row contains more complex concepts that decompose into three elements, where we visualize the removal of one element at a time.
Note that the results show non-trivial and profound links between concepts. For example, the \nl{snake} in~\cref{fig:single_image} is constructed as a \nl{twisted gecko}, the \nl{camel} borrows its skin texture and color from the \nl{cashmere}, \etc. These examples demonstrate associations beyond textual correlations, based on visual similarities such as shape, texture, and color. Additionally, note that the decompositions represent \emph{various different representation strategies}. Some involve a mixture of shape and appearance (\eg, \nl{snail}), others involve compositional features added gradually (\eg, \nl{impression of Japanese serenity}), \etc. Finally, observe that the \nl{painter} decomposition demonstrates reliance on renowned artistic styles, such that when \nl{Monet} is removed, the painting disappears, even though \nl{Monet} did not appear explicitly in the input prompt.~\cref{app:renown_styles} presents a further investigation of this reliance on artistic styles.

\noindent{\bf Coefficient manipulation\quad}
\cref{fig:manipulations} presents examples of coefficient manipulations.
First, we present examples of homograph concepts (first, second row of~\cref{fig:manipulations}). Observe that in some cases, \eg, \nl{big apple}, both meanings are generated separately in the image (an apple, New York City), while other cases, \eg, \nl{crane}, \nl{date} generate a single object. Even in the latter case, the manipulation shows that \emph{both meanings impact the generated image, implicitly}. For example, when reducing the element \nl{stork} from \nl{crane}, the structure of the crane changes. Evidently, the model employs both meanings simultaneously, borrowing the appearance from the machine and the shape from the bird. 

Next, we present examples of element inspection. In the first example, we visualize the impact of \nl{nerdy} on the concept \nl{professor}. As can be observed, it controls the professor's baldness, the glasses, the suit, and the library in the background. Secondly, we inspect exemplar interpolations. Observe that the element \nl{Trump} in an image of a \nl{president} controls semantic features borrowed from the identity of Donald Trump while removing this element results in a president that resembles Obama. This example directly demonstrates an interpolation of exemplars. Features from both Obama and Trump are employed \emph{simultaneously} in the image generation process. This phenomenon suggests that diffusion models can also memorize by mixing inputs, beyond exact single-sample reconstruction. Further investigation of exemplar-based representations can be found in~\cref{app:exemplars}.

\subsection{Quantitative Results}

\noindent{\bf Baselines\quad} As far as we can ascertain, our work is the first to tackle concept representations in text-to-image diffusion models.
We, therefore, compare our method with reasonable baselines and adaptations of existing methods. 
First, we consider a prompt tuning method, \emph{Hard Prompts Made Easy (PEZ)~\cite{Wen2023HardPM}}, which aims to learn a prompt that will reproduce an input set of training images.
Second, we consider two baselines that leverage
the state-of-the-art image captioning model BLIP-2~\cite{Li2023BLIP2BL}: (i) 
\emph{BLIP-2 sentence} extracts a single caption per concept by decoding the mean CLIP embedding of the training images. 
(ii) \emph{BLIP-2 token} creates one caption per image and constructs a single pseudo-token from the captions, where each token is weighted by its frequency in the captions. 
Finally, we consider CNN-based concept interpretability methods. As mentioned in \cref{sec:related}, these methods are not directly applicable to diffusion models. Therefore, we present comparisons to the closest adaptations of these methods. We decompose the activation of the UNet features from different denoising steps into concept activation vectors (CAVs) using \emph{k-means~\cite{Ghorbani2019TowardsAC}}, \emph{PCA~\cite{Zhang2020InvertibleCE}} and \emph{NMF~\cite{Fel2022CRAFTCR}}. At inference, we project the intermediate activations into the learned space, see \cref{app:concept_based_implementation} for more details.

\noindent{\bf Metrics\quad} For each concept, we test the \emph{faithfulness} and the \emph{diversity} of the decompositions. We use a test set of $100$ seeds to generate images with $w^c$ and with each method. 
Then, we employ three types of metrics: (i) \emph{Pairwise Similarity}, to measure the faithfulness of the decomposition w.r.t. each of the concept images.
We report the mean CLIP~\cite{radford2021learning} image similarity and the LPIPS~\cite{zhang2018unreasonable} score. 
(ii) \emph{Distribution similarity}, to measure the faithfulness of the decomposition w.r.t. the concept distribution. We report the FID~\cite{Heusel2017GANsTB} score with respect to the concept images for each concept separately (\emph{FID per concept}) and for all concepts as a dataset (\emph{FID entire set}).
(iii) We employ SentenceBERT~\cite{reimers-2019-sentence-bert} to measure the element diversity by estimating the dissimilarity of the tokens in the decomposition (\emph{Token diversity}). This metric further substantiates the \emph{meaningfulness} by showing that the decomposition is diverse.

\begin{figure*}[t]
    \centering
    \includegraphics[width=1.00\linewidth, clip]{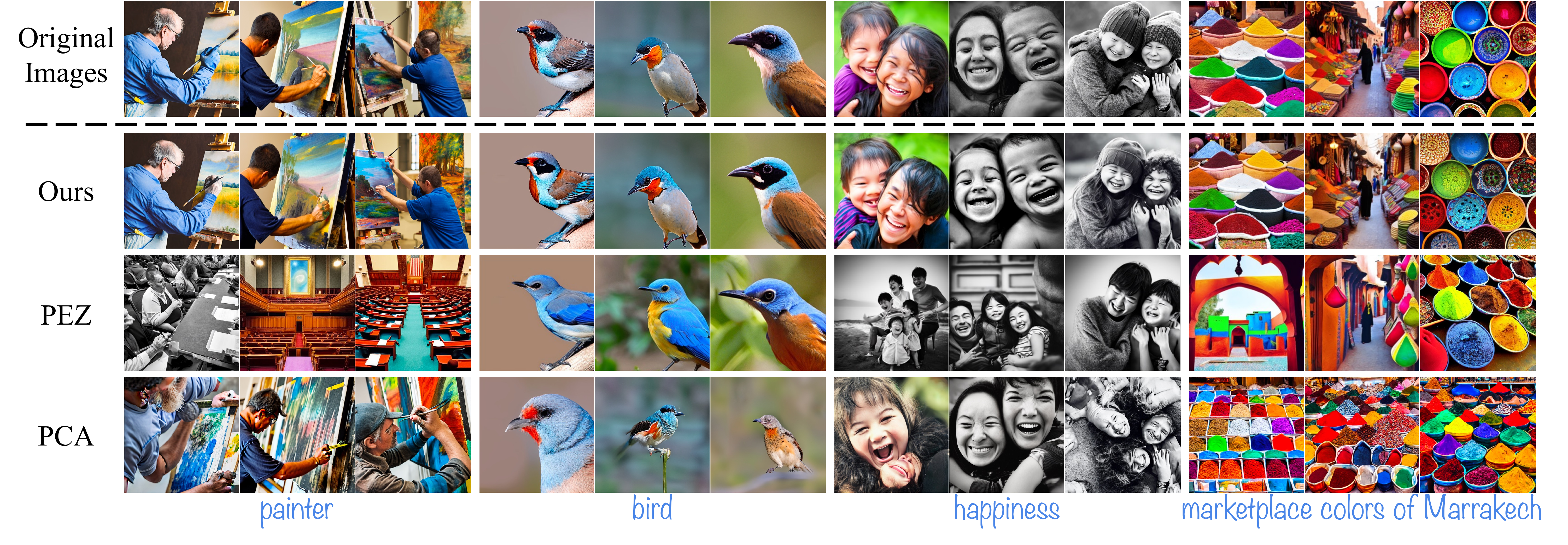}
    \vspace{-26px}
    \caption{ 
     Feature reconstruction comparison to the leading baselines. For each concept (column) we generate the images using the same random noise with our method and the leading baselines, and compare to the original concept images generated by Stable Diffusion (Original Images).  
    }
    \label{fig:reconstruction}
    \vspace{-14px}
\end{figure*}

\begin{table*}[t]
\small
\centering
\setlength{\tabcolsep}{4pt}
\addtolength{\belowcaptionskip}{-8pt}
\caption{Qualitative comparison of the decompositions by our method and the leading baselines.}
\vspace{-10px}
    \begin{tabular}{p{1.5cm}p{4cm}p{2.4cm}p{5.2cm}}
    \toprule
      Concept & PEZ & BLIP-2 sentence & \textbf{\textsc{Conceptor}} \\
      \midrule
         Rapper & \texttt{marin, prodigy, sturridge, noneeminem} & man, hat, shirt & \texttt{Tupac, Drake, Khalifa, Weekend, Khaled, hood, hat} \\  \midrule
          Medieval village life & \texttt{moderated, humpday, giftideas, shistory} &  people, medieval clothing, village & \texttt{caravan, peasant, medieval, countryside, farming} \\ \midrule
          Elegance on a plate & \texttt{silver, chesterfield, dinner, moschbizitalk} &  plate, vegetable, meat & \texttt{plating, chef, beauty, dessert, saucer, porcelain} \\
         \bottomrule
    \end{tabular}
    \label{tb:decompositions}
    \vspace{-5px}
\end{table*}
\noindent{\bf Results\quad} The results, averaged across all $188$ concepts, are reported in~\cref{tb:reconstruction}. As can be seen, our method outperforms all baselines across all faithfulness metrics. Notably, our LPIPS score is at least $10\%$ lower than that of all baselines, indicating that our method is faithful to the concept images. Importantly, both FID metrics obtained by our method are the lowest by a big margin (at least $20$ per concept and $8$ for the entire set). The scales of the FID scores are very different since the per concept FID was calculated on sets of merely $100$ images. However, we report both metrics for completeness.
Focusing on the more reliable FID score on the entire set, we calculate the  ``ground-truth'' FID between the train and test set and obtain a score of $7.1$, which is fairly close to our score. These results establish \textsc{Conceptor}'s ability to provide \emph{faithful} decompositions.
Considering the scores by SentenceBERT, \textsc{Conceptor} is only superseded by PEZ. However, PEZ produces a significant amount of uninterpretable tokens (see~\cref{tb:decompositions}). 
The matrix factorization methods do not produce text, therefore we cannot compute the SentenceBERT score. Instead, we enclose in~\cref{app:componenet} the top principal components learned for the concepts from~\cref{fig:reconstruction}. As can be seen, the obtained components do not appear to be coherent or interpretable, \ie, these methods violate the \emph{meaningfulness} criterion.

Next, we conduct qualitative comparisons between \textsc{Conceptor} and the leading baselines.~\cref{tb:decompositions} compares the textual decompositions, showing that \textsc{Conceptor} learns diverse and meaningful decompositions. Some concepts, such as \nl{a rapper} are dominated by exemplars (\eg \nl{Drake}, \nl{Tupac}), while others, such as \nl{Medival village life}, are a composition of semantically related concepts (\eg, \nl{caravan}, \nl{peasant}).
In contrast, the baselines either produce decompositions that are not interpretable, \ie, violate the \emph{meaningfulness} criterion ({PEZ}), or are oversimplistic ({BLIP-2}). Please refer to~\cref{app:complex_wordcloud} for word cloud visualizations of \textsc{Conceptor} over complex prompts.
\cref{fig:reconstruction} presents a visual comparison to the leading baselines given the same seeds. As can be observed, \textsc{Conceptor} successfully preserves the image features (\ie, upholds \emph{faithfulness}), even when the concept entails detailed features. For example, the \nl{painter} images demonstrate a \emph{reconstruction of the paintings}. Conversely, the baseline methods do not accurately embody all features of the concept.

\noindent{\bf User study\quad} To further substantiate our \emph{meaningfulness} and \emph{humans' ability} to understand our decompositions, we conduct a user study. In the study, we randomly draw $8$ concepts, $2$ from each of our data categories: (a) professions, (b) abstract concepts, (c) basic concepts (ConceptNet,  CIFAR-10), and (d) complex concepts. For each concept, the users were presented with $6$ random concept images and $3$ possible textual decompositions, as provided by \textsc{Conceptor}, and the two leading baselines that extract text (\emph{BLIP-2 sentence} and \emph{PEZ}). The users were asked to select the decomposition that best captures all the features presented in the images. Overall, we collected $160$ responses for all questions. Of those $160$ responses, $70\%$ favored the decomposition by \textsc{Conceptor} above all alternatives, $26.9\%$ favored the captioning by BLIP-2, and $3.1\%$ selected PEZ, further establishing \textsc{Conceptor}’s meaningfulness, even compared to the natural captioning alternative.

\noindent{\bf Robustness experiments\quad}
In~\cref{app:robustness} we conduct extensive \emph{robustness} tests, to demonstrate two important properties: (i) \textsc{Conceptor} is robust to \emph{different choices of the training set and initialization}, and (ii) \textsc{Conceptor} generalizes the training task to \emph{test images}. Experiment (i) employs $3$ different training sets and initializations and tests the \emph{exact} match of tokens between the $3$ resulting decompositions. The results demonstrate that the decomposition is consistent across all sets- $72-80\%$ of the top-$10$ elements, and $63-70\%$ of the top-$25$ elements are preserved across all choices. Experiment (ii) shows that $w^*$ is able to denoise \emph{test} concept images from \emph{any denoising step}, thus $w^*$ indeed captures the concept features beyond the selection of training images.

In conclusion, we find that the baselines either violate the \emph{meaningfulness} criterion, \ie, produce uninterpretable decompositions (\emph{k-means, PCA, NMF, PEZ}, see~\cref{tb:decompositions,app:componenet}, user study) or the \emph{faithfulness} criterion (\emph{BLIP-2}, see poor performance in~\cref{tb:reconstruction}). We note that for both criteria and in all experiments, \textsc{Conceptor} significantly outperforms all baselines and obtains \emph{robust}, \emph{faithful}, and \emph{meaningful} decompositions (see~\cref{tb:reconstruction,tb:decompositions,fig:reconstruction,app:robustness}, user study).

\begin{table*}[t]
\setlength{\belowcaptionskip}{-4pt}
\small
\centering
\setlength{\tabcolsep}{2pt}
\caption{Ablation study of our method, conducted on the professions subset~\cite{1901.09451}. 
} 
\vspace{-10px}
\begin{tabular}{l@{\hspace{0.4cm}} c@{\hspace{0.4cm}} c@{\hspace{0.4cm}} c@{\hspace{0.3cm}} c c} 
\toprule
 Method &  {CLIP pairwise$\uparrow$}& LPIPS$\downarrow$   & FID per concept$\downarrow$ & & \begin{tabular}{c} Token diversity$\uparrow$ \end{tabular}  \\
\midrule
\begin{tabular}{l} \textsc{Conceptor} \end{tabular}  & \textbf{87.0} $\pm$ 5.5 & \textbf{0.45} $\pm$ 0.07 & \textbf{107.96} $\pm$ 31.0  &  & {69.7} $\pm$ 3.4 \\
\begin{tabular}{l} w/o MLP  \end{tabular}  & {78.0} $\pm$ 6.7 & {0.55} $\pm$ 0.06 & {142.88} $\pm$ 45.1  &  & \textbf{75.9} $\pm$ 3.0 \\
\begin{tabular}{l} w/o~\cref{eq:sparsity}  \end{tabular}  & {80.3} $\pm$ 11.6 & {0.52} $\pm$ 0.09 & {146.4} $\pm$ 63.4  &  & {73.2} $\pm$ 2.1 \\
\begin{tabular}{l} $n=10$ \end{tabular}  & {82.9} $\pm$ 7.8 & {0.49} $\pm$ 0.11 & {129.41} $\pm$ 55.3  & & {54.6} $\pm$ 9.4 \\
\begin{tabular}{l} $n=100$ \end{tabular}  & {85.6} $\pm$ 6.9 & {0.47} $\pm$ 0.07 & {114.36} $\pm$ 39.7  && {72.8} $\pm$ 1.8 \\
\begin{tabular}{l} CLIP top words\end{tabular}  &  80.1 $\pm$ 9.9 & 0.513 $\pm$ 0.1  & 130.9 $\pm$ 57.2 & & 66.3 $\pm$ 3.9 \\
\bottomrule 
\end{tabular}
\label{tb:ablation}
\vspace{-10px}
\end{table*}
\subsubsection{Ablation Study}
We conduct an ablation study to examine the impact of each component on our method. First, we ablate the choice of employing an MLP to learn the coefficients and instead learn them directly. Next, we ablate our sparsity loss and the choice of $n=50$. Last, we ablate our choice of vocabulary $\mathcal{V}$ and instead extract the top $50$ tokens by their CLIP similarity to the mean image. 

The results are listed in~\cref{tb:ablation}. Replacing the MLP with a vector of weights is \emph{detrimental to all faithfulness metrics}. This demonstrates the importance of our learned MLP, as it leverages the rich semantic information learned by CLIP, rather than optimizing a huge set of coefficients without any semantic understanding.
Without the sparsity loss (\cref{eq:sparsity}), the top $50$ tokens do not necessarily reflect the learned token $w^*_N$, and all metrics except for token diversity deteriorate. 
Additionally, observe that the performance decreases when employing $n=10$ since the decomposition is not rich enough to represent all features. For $n=100$, the results are similar to the full method, other than the diversity which improves a little. This indicates that \textsc{Conceptor} is relatively stable to this parameter. 
Finally, when only considering the top words by CLIP similarity to the images, the performance decreases substantially, supporting the reliance of our method on a wide variety of tokens from the vocabulary, beyond the ones most correlated with the images in terms of textual semantics.

  \begin{table}[t]
\small
\centering
\setlength{\tabcolsep}{2pt}
\setlength{\baselineskip}{-12pt}
\caption{Decomposition elements obtained by \textsc{Conceptor} that reveal potential biases.
} 
\vspace{-8px}
\begin{tabular}{l@{\hspace{0.4cm}} l } 
\toprule
    Concept  &  Decomposition  \\
\midrule
Secretary & \texttt{clerk}, \texttt{ wife}, \texttt{ womens}, \texttt{ girl}, \texttt{ ladies}... \\
Opera singer & \texttt{ obese}, \texttt{ soprano}, \texttt{ overweight}, \texttt{ fat}... \\
Pastor & \texttt{Nigerian}, \texttt{ gospel}, \texttt{ worship}... \\
Journalist & \texttt{stranger}, \texttt{ refugee}, \texttt{ jews}, \texttt{ tripod}... \\
Drinking & \texttt{cheating}, \texttt{ millennials}, \texttt{ blonde}... \\
\bottomrule 
\end{tabular}
\label{tb:biases}
\vspace{-12px}
\end{table}

\subsection{Bias Detection}
An important capability of our method is bias discovery. Text-to-image models have been shown to represent social biases~\cite{luccioni2023stable}. The decompositions obtained by \textsc{Conceptor} can be used to discover such biases by analyzing the decomposition elements. \cref{tb:biases} lists some concepts that expose social insensitivities. Note that our method detects behaviors that are not necessarily observable visually such as a connection between \nl{Jews} and \nl{journalists}. These findings substantiate the need to conduct more research on concept representations in text-to-image models, as biases can impact the generation even if they are hard to detect visually. Using our method, users can also choose to generate debiased versions of these concepts by decreasing the coefficients corresponding to the biased tokens. More details on such manipulations can be found in~\cref{app:debiasing}.

\begin{figure}[t]
    \centering
    \includegraphics[width=1\linewidth, clip]{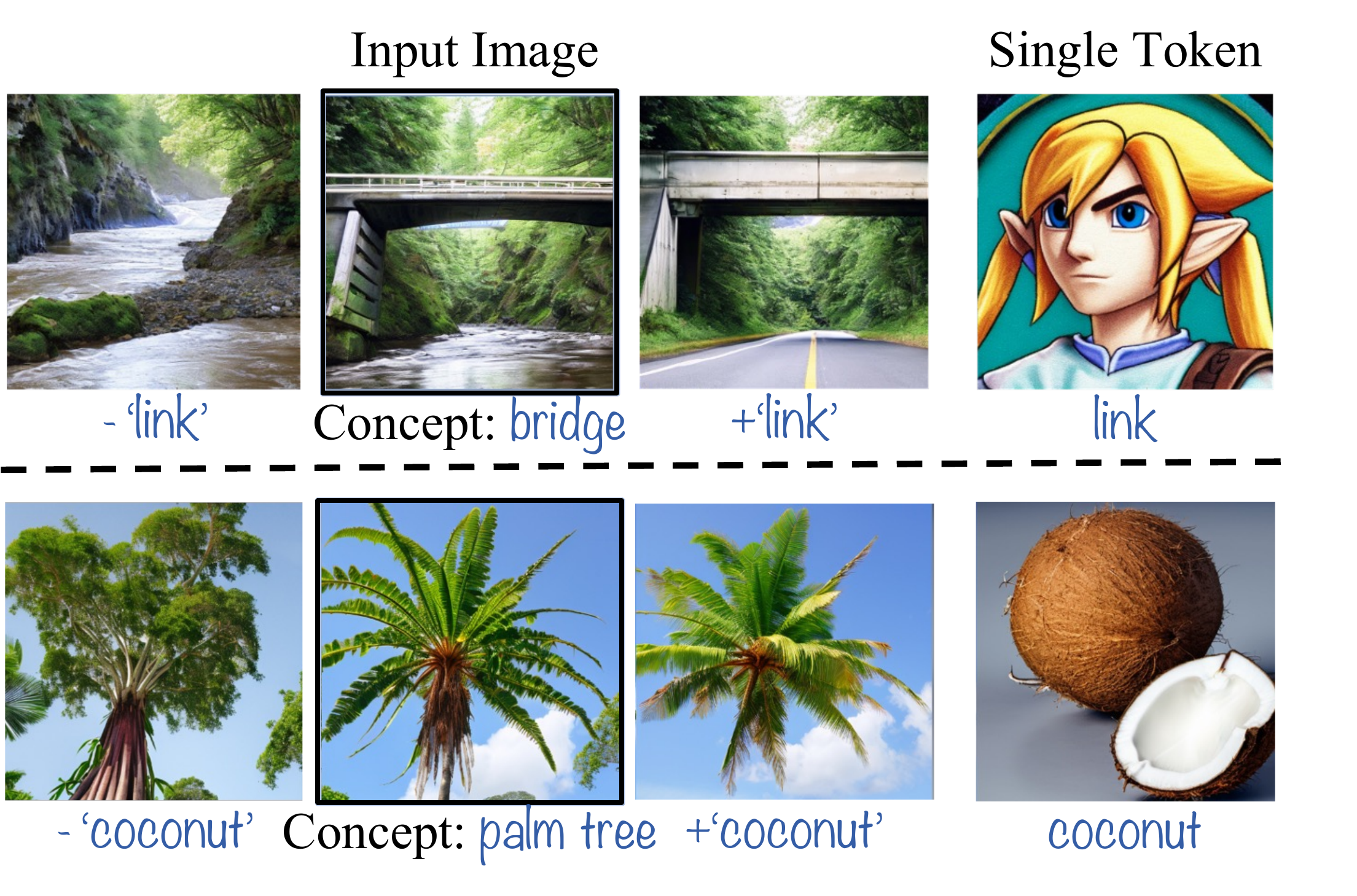}
    \vspace{-24px}
    \caption{
    Limitations. Examples of elements that impact the generated image differently from their single-token meaning. On its own, each token generates an image that is completely different than the input image (Single Token). Given the context of the decomposition, the effect of the token changes.
    }
    \vspace{-16px}
    \label{fig:limitations}
\end{figure}

\section{Discussion and Limitations}

While our method provides faithful and interpretable concept decompositions, there are some limitations to consider. First, we find that the visual impact of an element is not always completely aligned with its impact as a single token, \ie, \emph{the impact of each token depends on the context of the decomposition}.~\cref{fig:limitations} demonstrates such cases. For each concept, we visualize the effect of manipulating the token, and the result of keeping only the token of interest (Single Token).
Note that the influence of a token on the generated image differs from its influence as a sole token. For example, the token \nl{link} on its own produces an image of the video game character, Link. However, in the context of a \nl{bridge}, it adds a solid bridge (a \emph{link}) between the two edges of the image. 

Second, our method is limited to elements that are single tokens, therefore a complex phrase (\eg, \nl{palm tree}) will not be included as a single element in our decomposition. However, as mentioned, our construction of the linear combination mitigates the impact of this limitation. Since the context of the decomposition can change the meaning of an element, complex relations can be formed between single tokens by leveraging the context. This is exemplified by our ability to decompose complex concepts, that require hierarchical reasoning (\eg, \nl{elegance on a plate}, \nl{rainbow dewdrops}).

\section{Conclusions}
How does a generative model perceive the world? Focusing on text-to-image diffusion models, we investigate the model's internal knowledge of real-world concepts.
We present \textsc{Conceptor}, a method to provide a human-understandable decomposition for a textual concept. Through extensive experiments, we show that \textsc{Conceptor} provides interpretations that are meaningful, robust, and faithful to the concept representation by the model.
Using \textsc{Conceptor}, we obtain various interesting observations on the learned concept representations. Via a per-image decomposition scheme, we observe non-trivial connections between concepts in ways that transcend the lexical meaning of the tokens. 
Furthermore, our method exposes less intuitive behaviors such as the reliance on exemplars, mixing dual meanings of concepts, or non-trivial biases. In all cases, the novel paradigm allows us to shed new light on a model that, similar to other foundation models, can still be considered an enigma.

{\small
\bibliographystyle{ieee_fullname}
\bibliography{egbib}

\begin{thebibliography}{10}\itemsep=-1pt

\bibitem{Agarap2018DeepLU}
Abien~Fred Agarap.
\newblock Deep learning using rectified linear units (relu).
\newblock {\em ArXiv}, abs/1803.08375, 2018.

\bibitem{Balaji2022eDiffITD}
Yogesh Balaji, Seungjun Nah, Xun Huang, Arash Vahdat, Jiaming Song, Karsten Kreis, Miika Aittala, Timo Aila, Samuli Laine, Bryan Catanzaro, Tero Karras, and Ming-Yu Liu.
\newblock ediff-i: Text-to-image diffusion models with an ensemble of expert denoisers.
\newblock {\em ArXiv}, abs/2211.01324, 2022.

\bibitem{Carlini2023ExtractingTD}
Nicholas Carlini, Jamie Hayes, Milad Nasr, Matthew Jagielski, Vikash Sehwag, Florian Tram{\`e}r, Borja Balle, Daphne Ippolito, and Eric Wallace.
\newblock Extracting training data from diffusion models.
\newblock {\em ArXiv}, abs/2301.13188, 2023.

\bibitem{chefer2023attend}
Hila Chefer, Yuval Alaluf, Yael Vinker, Lior Wolf, and Daniel Cohen-Or.
\newblock Attend-and-excite: Attention-based semantic guidance for text-to-image diffusion models, 2023.

\bibitem{Chefer_2021_ICCV}
Hila Chefer, Shir Gur, and Lior Wolf.
\newblock Generic attention-model explainability for interpreting bi-modal and encoder-decoder transformers.
\newblock In {\em Proceedings of the IEEE/CVF International Conference on Computer Vision (ICCV)}, pages 397--406, October 2021.

\bibitem{Chefer_2021_CVPR}
Hila Chefer, Shir Gur, and Lior Wolf.
\newblock Transformer interpretability beyond attention visualization.
\newblock In {\em Proceedings of the IEEE/CVF Conference on Computer Vision and Pattern Recognition (CVPR)}, pages 782--791, June 2021.

\bibitem{chuang2023debiasing}
Ching-Yao Chuang, Varun Jampani, Yuanzhen Li, Antonio Torralba, and Stefanie Jegelka.
\newblock Debiasing vision-language models via biased prompts.
\newblock {\em arXiv preprint arXiv:2302.00070}, 2023.

\bibitem{1901.09451}
Maria De-Arteaga, Alexey Romanov, Hanna Wallach, Jennifer Chayes, Christian Borgs, Alexandra Chouldechova, Sahin Geyik, Krishnaram Kenthapadi, and Adam~Tauman Kalai.
\newblock Bias in bios: A case study of semantic representation bias in a high-stakes setting.
\newblock 2019.

\bibitem{Fel2023Holistic}
Thomas Fel, Victor Boutin, Mazda Moayeri, R{\'{e}}mi Cad{\`{e}}ne, Louis B{\'{e}}thune, L{\'{e}}o And{\'{e}}ol, Mathieu Chalvidal, and Thomas Serre.
\newblock A holistic approach to unifying automatic concept extraction and concept importance estimation.
\newblock {\em CoRR}, abs/2306.07304, 2023.

\bibitem{Fel2022CRAFTCR}
Thomas Fel, Agustin Picard, Louis B{\'e}thune, Thibaut Boissin, David Vigouroux, Julien Colin, R'emi Cadene, and Thomas Serre.
\newblock Craft: Concept recursive activation factorization for explainability.
\newblock {\em 2023 IEEE/CVF Conference on Computer Vision and Pattern Recognition (CVPR)}, pages 2711--2721, 2022.

\bibitem{gafni2022make}
Oran Gafni, Adam Polyak, Oron Ashual, Shelly Sheynin, Devi Parikh, and Yaniv Taigman.
\newblock Make-a-scene: Scene-based text-to-image generation with human priors.
\newblock {\em arXiv preprint arXiv:2203.13131}, 2022.

\bibitem{gal2022image}
Rinon Gal, Yuval Alaluf, Yuval Atzmon, Or Patashnik, Amit~H Bermano, Gal Chechik, and Daniel Cohen-Or.
\newblock An image is worth one word: Personalizing text-to-image generation using textual inversion.
\newblock {\em arXiv preprint arXiv:2208.01618}, 2022.

\bibitem{gandikota2023unified}
Rohit Gandikota, Hadas Orgad, Yonatan Belinkov, Joanna Materzy\'nska, and David Bau.
\newblock Unified concept editing in diffusion models.
\newblock {\em arXiv preprint arXiv:2308.14761}, 2023.

\bibitem{geva2023dissecting}
Mor Geva, Jasmijn Bastings, Katja Filippova, and Amir Globerson.
\newblock Dissecting recall of factual associations in auto-regressive language models.
\newblock {\em arXiv preprint arXiv:2304.14767}, 2023.

\bibitem{geva-etal-2022-transformer}
Mor Geva, Avi Caciularu, Kevin Wang, and Yoav Goldberg.
\newblock Transformer feed-forward layers build predictions by promoting concepts in the vocabulary space.
\newblock In {\em Proceedings of the 2022 Conference on Empirical Methods in Natural Language Processing}, pages 30--45, Abu Dhabi, United Arab Emirates, Dec. 2022. Association for Computational Linguistics.

\bibitem{Ghorbani2019TowardsAC}
Amirata Ghorbani, James Wexler, James~Y. Zou, and Been Kim.
\newblock Towards automatic concept-based explanations.
\newblock In {\em Neural Information Processing Systems}, 2019.

\bibitem{han2023svdiff}
Ligong Han, Yinxiao Li, Han Zhang, Peyman Milanfar, Dimitris Metaxas, and Feng Yang.
\newblock Svdiff: Compact parameter space for diffusion fine-tuning.
\newblock {\em arXiv preprint arXiv:2303.11305}, 2023.

\bibitem{hertz2022prompt}
Amir Hertz, Ron Mokady, Jay Tenenbaum, Kfir Aberman, Yael Pritch, and Daniel Cohen-Or.
\newblock Prompt-to-prompt image editing with cross attention control.
\newblock {\em arXiv preprint arXiv:2208.01626}, 2022.

\bibitem{Heusel2017GANsTB}
Martin Heusel, Hubert Ramsauer, Thomas Unterthiner, Bernhard Nessler, and Sepp Hochreiter.
\newblock Gans trained by a two time-scale update rule converge to a local nash equilibrium.
\newblock In {\em NIPS}, 2017.

\bibitem{ho2020denoising}
Jonathan Ho, Ajay Jain, and Pieter Abbeel.
\newblock Denoising diffusion probabilistic models.
\newblock {\em Advances in Neural Information Processing Systems}, 33:6840--6851, 2020.

\bibitem{Kim2017InterpretabilityBF}
Been Kim, Martin Wattenberg, Justin Gilmer, Carrie~J. Cai, James Wexler, Fernanda~B. Vi{\'e}gas, and Rory Sayres.
\newblock Interpretability beyond feature attribution: Quantitative testing with concept activation vectors (tcav).
\newblock In {\em International Conference on Machine Learning}, 2017.

\bibitem{Krizhevsky2009LearningML}
Alex Krizhevsky.
\newblock Learning multiple layers of features from tiny images.
\newblock 2009.

\bibitem{kumari2022multi}
Nupur Kumari, Bingliang Zhang, Richard Zhang, Eli Shechtman, and Jun-Yan Zhu.
\newblock Multi-concept customization of text-to-image diffusion.
\newblock {\em arXiv preprint arXiv:2212.04488}, 2022.

\bibitem{li-etal-2021-implicit}
Belinda~Z. Li, Maxwell Nye, and Jacob Andreas.
\newblock Implicit representations of meaning in neural language models.
\newblock In {\em Proceedings of the 59th Annual Meeting of the Association for Computational Linguistics and the 11th International Joint Conference on Natural Language Processing (Volume 1: Long Papers)}, pages 1813--1827, Online, Aug. 2021. Association for Computational Linguistics.

\bibitem{Li2023BLIP2BL}
Junnan Li, Dongxu Li, Silvio Savarese, and Steven Hoi.
\newblock Blip-2: Bootstrapping language-image pre-training with frozen image encoders and large language models.
\newblock {\em ArXiv}, abs/2301.12597, 2023.

\bibitem{li2023emergent}
Kenneth Li, Aspen~K Hopkins, David Bau, Fernanda Vi{\'e}gas, Hanspeter Pfister, and Martin Wattenberg.
\newblock Emergent world representations: Exploring a sequence model trained on a synthetic task.
\newblock In {\em The Eleventh International Conference on Learning Representations}, 2023.

\bibitem{lovering-pavlick-2022-unit}
Charles Lovering and Ellie Pavlick.
\newblock Unit testing for concepts in neural networks.
\newblock {\em Transactions of the Association for Computational Linguistics}, 10:1193--1208, 2022.

\bibitem{luccioni2023stable}
Alexandra~Sasha Luccioni, Christopher Akiki, Margaret Mitchell, and Yacine Jernite.
\newblock Stable bias: Analyzing societal representations in diffusion models.
\newblock {\em arXiv preprint arXiv:2303.11408}, 2023.

\bibitem{nichol2021glide}
Alex Nichol, Prafulla Dhariwal, Aditya Ramesh, Pranav Shyam, Pamela Mishkin, Bob McGrew, Ilya Sutskever, and Mark Chen.
\newblock Glide: Towards photorealistic image generation and editing with text-guided diffusion models.
\newblock {\em arXiv preprint arXiv:2112.10741}, 2021.

\bibitem{patel2022mapping}
Roma Patel and Ellie Pavlick.
\newblock Mapping language models to grounded conceptual spaces.
\newblock In {\em International Conference on Learning Representations}, 2022.

\bibitem{radford2021learning}
Alec Radford, Jong~Wook Kim, Chris Hallacy, Aditya Ramesh, Gabriel Goh, Sandhini Agarwal, Girish Sastry, Amanda Askell, Pamela Mishkin, Jack Clark, et~al.
\newblock Learning transferable visual models from natural language supervision.
\newblock In {\em International Conference on Machine Learning}, pages 8748--8763. PMLR, 2021.

\bibitem{ram2022you}
Ori Ram, Liat Bezalel, Adi Zicher, Yonatan Belinkov, Jonathan Berant, and Amir Globerson.
\newblock What are you token about? dense retrieval as distributions over the vocabulary.
\newblock {\em arXiv preprint arXiv:2212.10380}, 2022.

\bibitem{Ramesh2022Hierarchical}
Aditya Ramesh, Prafulla Dhariwal, Alex Nichol, Casey Chu, and Mark Chen.
\newblock Hierarchical text-conditional image generation with clip latents.
\newblock {\em ArXiv}, 2022.

\bibitem{ramesh2021zero}
Aditya Ramesh, Mikhail Pavlov, Gabriel Goh, Scott Gray, Chelsea Voss, Alec Radford, Mark Chen, and Ilya Sutskever.
\newblock Zero-shot text-to-image generation.
\newblock In {\em International Conference on Machine Learning}, pages 8821--8831. PMLR, 2021.

\bibitem{rassin2022dalle}
Royi Rassin, Shauli Ravfogel, and Yoav Goldberg.
\newblock Dalle-2 is seeing double: Flaws in word-to-concept mapping in text2image models, 2022.

\bibitem{reimers-2019-sentence-bert}
Nils Reimers and Iryna Gurevych.
\newblock Sentence-bert: Sentence embeddings using siamese bert-networks.
\newblock In {\em Proceedings of the 2019 Conference on Empirical Methods in Natural Language Processing}. Association for Computational Linguistics, 11 2019.

\bibitem{rombach2022high}
Robin Rombach, Andreas Blattmann, Dominik Lorenz, Patrick Esser, and Bj{\"o}rn Ommer.
\newblock High-resolution image synthesis with latent diffusion models.
\newblock In {\em Proceedings of the IEEE/CVF Conference on Computer Vision and Pattern Recognition}, pages 10684--10695, 2022.

\bibitem{ronneberger2015u}
Olaf Ronneberger, Philipp Fischer, and Thomas Brox.
\newblock U-net: Convolutional networks for biomedical image segmentation.
\newblock In {\em International Conference on Medical image computing and computer-assisted intervention}, pages 234--241. Springer, 2015.

\bibitem{ruiz2022dreambooth}
Nataniel Ruiz, Yuanzhen Li, Varun Jampani, Yael Pritch, Michael Rubinstein, and Kfir Aberman.
\newblock Dreambooth: Fine tuning text-to-image diffusion models for subject-driven generation.
\newblock {\em arXiv preprint arXiv:2208.12242}, 2022.

\bibitem{Saharia2022PhotorealisticTD}
Chitwan Saharia, William Chan, Saurabh Saxena, Lala Li, Jay Whang, Emily~L. Denton, Seyed Kamyar~Seyed Ghasemipour, Burcu~Karagol Ayan, Seyedeh~Sara Mahdavi, Raphael~Gontijo Lopes, Tim Salimans, Jonathan Ho, David~J. Fleet, and Mohammad Norouzi.
\newblock Photorealistic text-to-image diffusion models with deep language understanding.
\newblock {\em ArXiv}, abs/2205.11487, 2022.

\bibitem{sohl2015deep}
Jascha Sohl-Dickstein, Eric Weiss, Niru Maheswaranathan, and Surya Ganguli.
\newblock Deep unsupervised learning using nonequilibrium thermodynamics.
\newblock In {\em International Conference on Machine Learning}, pages 2256--2265. PMLR, 2015.

\bibitem{somepalli2022diffusion}
Gowthami Somepalli, Vasu Singla, Micah Goldblum, Jonas Geiping, and Tom Goldstein.
\newblock Diffusion art or digital forgery? investigating data replication in diffusion models.
\newblock {\em arXiv preprint arXiv:2212.03860}, 2022.

\bibitem{Speer2013ConceptNet5A}
Robyn Speer and Catherine Havasi.
\newblock Conceptnet 5: A large semantic network for relational knowledge.
\newblock In {\em The People's Web Meets NLP}, 2013.

\bibitem{Voyov2023Pplus}
Andrey Voynov, Qinghao Chu, Daniel Cohen-Or, and Kfir Aberman.
\newblock P+: Extended textual conditioning in text-to-image generation, 2023.

\bibitem{Wen2023HardPM}
Yuxin Wen, Neel Jain, John Kirchenbauer, Micah Goldblum, Jonas Geiping, and Tom Goldstein.
\newblock Hard prompts made easy: Gradient-based discrete optimization for prompt tuning and discovery.
\newblock {\em ArXiv}, abs/2302.03668, 2023.

\bibitem{yu2022scaling}
Jiahui Yu, Yuanzhong Xu, Jing~Yu Koh, Thang Luong, Gunjan Baid, Zirui Wang, Vijay Vasudevan, Alexander Ku, Yinfei Yang, Burcu~Karagol Ayan, et~al.
\newblock Scaling autoregressive models for content-rich text-to-image generation.
\newblock {\em arXiv preprint arXiv:2206.10789}, 2022.

\bibitem{yun2023visionlanguage}
Tian Yun, Usha Bhalla, Ellie Pavlick, and Chen Sun.
\newblock Do vision-language pretrained models learn composable primitive concepts?, 2023.

\bibitem{zhang2018unreasonable}
Richard Zhang, Phillip Isola, Alexei~A Efros, Eli Shechtman, and Oliver Wang.
\newblock The unreasonable effectiveness of deep features as a perceptual metric.
\newblock In {\em Proceedings of the IEEE conference on computer vision and pattern recognition}, pages 586--595, 2018.

\bibitem{Zhang2020InvertibleCE}
Ruihan Zhang, Prashan Madumal, Tim Miller, Krista~A. Ehinger, and Benjamin I.~P. Rubinstein.
\newblock Invertible concept-based explanations for cnn models with non-negative concept activation vectors.
\newblock In {\em AAAI Conference on Artificial Intelligence}, 2020.

\end{thebibliography}
}

\clearpage
\onecolumn
\appendix
\appendix

\section{Implementation Details}
\label{app:implementation}
\begin{figure*}[h]
    \centering
    \setlength{\belowcaptionskip}{-0pt}\includegraphics[width=1\linewidth, clip]{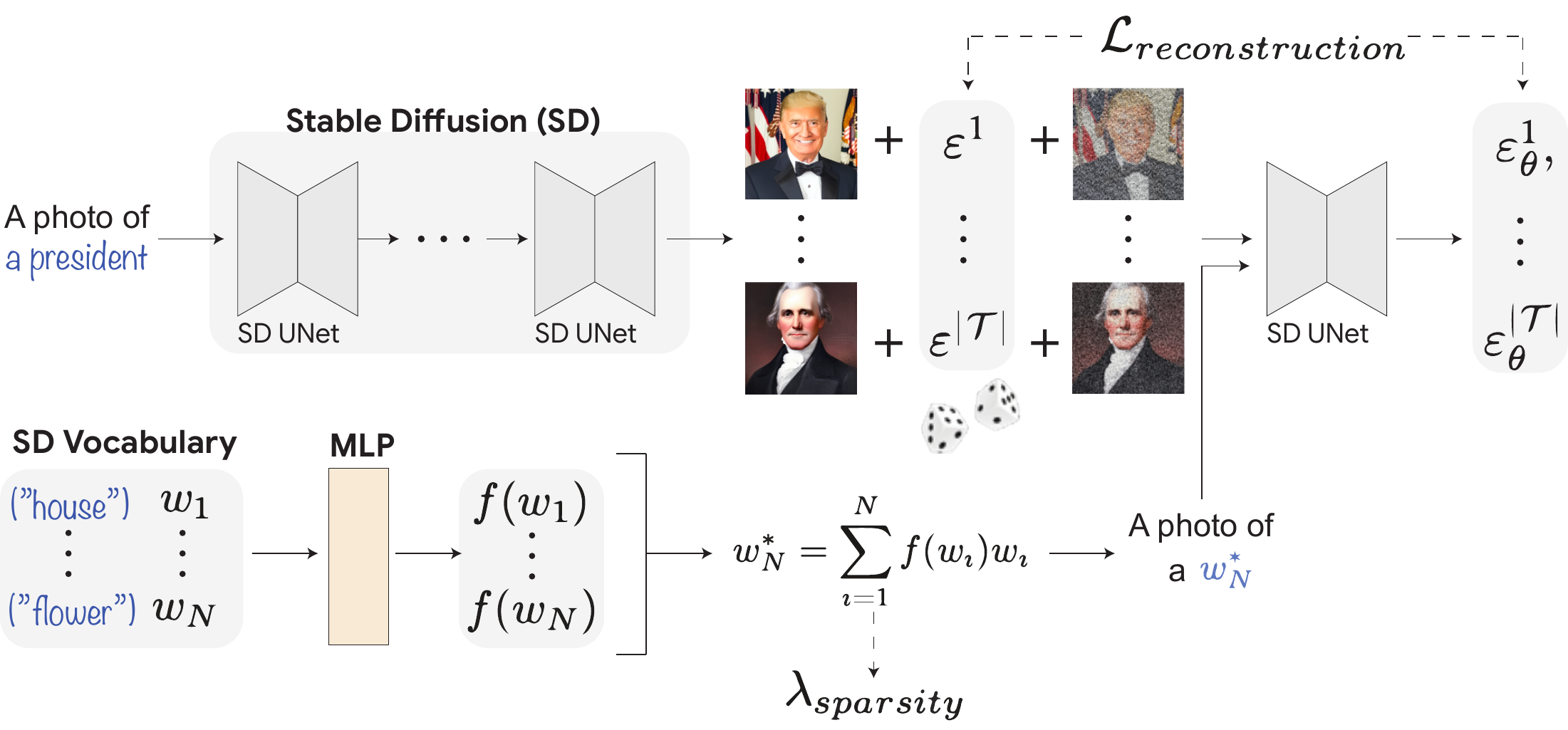}
    \caption{
     Illustration of the \textsc{Conceptor} method. Given the concept of interest (\eg, \nl{a president}), we generate $100$ concept images. Next, a learned MLP network maps each word embedding $w_i$ to a coefficient $f(w_i)$, and the pseudo token $w_N^*$ is constructed as a linear combination of the vocabulary. We then add random noises $\varepsilon^1,\dots,\varepsilon^{|\mathcal{T}|}$ to the images, and use the model to predict the noise based on the text \nl{a photo of a <$w_N^*$>}. We train the MLP with the objective of reconstructing the images ($\mathcal{L}_{reconstruction}$) and add a sparsity loss to encourage sparse coefficients ($\mathcal{L}_{sparsity}$).
    }
    \label{fig:method}
    \vspace{-0px}
\end{figure*}
\cref{fig:method} describes the pipeline of our method. Given an input concept (\eg, \nl{a president}), we obtain a set of representative concept images. At each optimization iteration, we add random noises to each of the images individually. Our learned MLP operates over the model's vocabulary to produce a coefficient for each of the word embeddings, and the resulting pseudo-token is used to denoise the training images. We employ a reconstruction loss to encourage the pseudo-token to contain the concept elements learned by the model and add a sparsity loss to maintain an interpretable and stable decomposition.

All of our experiments were conducted using a single A100 GPU with $40$GB of memory. We train our MLP as specified in~\cref{sec:method} of the main paper with $100$ images generated from the concept using seed $1024$ for a maximum of $500$ training steps with a batch size of $6$ (which is the largest batch size that could fit on our GPU). Additionally, we use a learning rate of $1e-3$ (grid searched on $5$ concepts between $1e-2, 1e-3, 1e-4$).
We conduct validation every $50$ optimization steps on $20$ images with a validation seed and select the iteration with the best CLIP pairwise similarity between the reconstruction and the concept images. 
We use the latest Stable Diffusion v2.1\footnote{\url{https://github.com/Stability-AI/stablediffusion}} text-to-image model employing the pre-trained text encoder from the OpenCLIP ViT-H model\footnote{\url{https://github.com/mlfoundations/open\_clip}}, with a fixed guidance scale of $7.5$.

Additionally, to filter out meaningless tokens such as punctuation marks we consider the vocabulary to be the top $5,000$ tokens by their CLIP similarity to the mean training image. We note that this filtering method is fairly coarse, and meaningless tokens remain in the vocabulary, however, we find that this choice improves the convergence time of our MLP such that $500$ iterations are enough to obtain meaningful decompositions. This choice is ablated in the main paper (see the CLIP top words ablation).
Our single-image decomposition scheme employs a CLIP ViT-B/32 model~\cite{radford2021learning}.
Please find our code attached as a ZIP file to reproduce our results.

\subsection{Concept-based baselines implementation details}
\label{app:concept_based_implementation}
To implement the concept-based baseline approaches, we follow the formulation in~\cite{Fel2023Holistic}.
During the generation of the training concept images, we collect the intermediate representations from the UNet at different denoising timesteps. Then, we decompose those representations using PCA, following~\cite{Zhang2020InvertibleCE}, NMF, following~\cite{Fel2022CRAFTCR} or k-means, following~\cite{Ghorbani2019TowardsAC}. We use a total of $n=50$ components, to match our decomposition size. Finally, at inference, we generate images from the concept using the test seed, $2$, while projecting and reconstructing the aforementioned intermediate representations both before passing them to the next block and to the upsampling (decoding) part of the UNet (via the skip-connections). At training and inference when using NMF we drop the negative activations. At inference, when using k-means, we employ the closest cluster as a reconstruction of an intermediate representation.

\begin{itemize} [leftmargin=*,topsep=0pt,itemsep=0pt,parsep=0pt]
    \item We compute PCA using $5$ iterations of randomized SVD solver with $10$ oversamples. 

    \item We compute k-means using $10$ k-means++ initialization and $100$ iterations of the Lloyd algorithm. 

    \item We compute NMF using NNDSVD initialization and run the Coordinate Descent solver for $200$ iterations.
 \end{itemize}

\section{Dataset}
\label{app:data}
In the following, we enclose the full list of concepts in each of our data subsets.

\paragraph{Bias in Bios concepts}
\nl{professor},
\nl{physician},
\nl{attorney},
\nl{photographer},
\nl{journalist},
\nl{nurse},
\nl{psychologist},
\nl{teacher},
\nl{dentist},
\nl{surgeon},
\nl{architect},
\nl{painter},
\nl{model},
\nl{poet},
\nl{filmmaker},
\nl{software engineer},
\nl{accountant},
\nl{composer},
\nl{dietitian},
\nl{comedian},
\nl{chiropractor},
\nl{pastor},
\nl{paralegal},
\nl{yoga teacher},
\nl{dj},
\nl{interior designer},
\nl{personal trainer},
\nl{rapper}

\paragraph{CIFAR-10 concepts}
\nl{airplane},  \nl{automobile},  \nl{bird},  \nl{cat},  \nl{deer},  \nl{dog},  \nl{frog},  \nl{horse},  \nl{ship},  \nl{truck}

\paragraph{Emotions}
\nl{affection},  \nl{anger}, \nl{disgust},  \nl{fear}, \nl{happiness},
\nl{honor},\nl{joy}, \nl{justice},  \nl{sadness}, \nl{beauty}

\paragraph{Actions}
\nl{clapping},  \nl{climbing},  \nl{drinking},  \nl{hugging},\nl{jumping}, \nl{pouring},  \nl{running},  \nl{sitting}, \nl{throwing},  \nl{walking}

\paragraph{Complex concepts}
\nl{Rainy New York Nights},
\nl{The Fashion of Abandoned Places},
\nl{Rainbow Dewdrops},
\nl{Aerial Autumn River},
\nl{Skateboarder’s Urban Flight},
\nl{Dive into Coral Reefs},
\nl{Vintage European Transit},
\nl{Star Trails over Mountains},
\nl{Marketplace Colors of Marrakesh},
\nl{Elegance on a Plate},
\nl{The Renaissance Astronaut},
\nl{The Surreal Floating Island},
\nl{Impression of Japanese Serenity},
\nl{Jazz in Abstract Colors},
\nl{The Confluence of Pop Art},
\nl{The Robotic Baroque Battle},
\nl{Cubist Bustling Market},
\nl{The Romantic Stormy Voyage},
\nl{The Botanist in Art Nouveau},
\nl{The Gothic Moonlit Castle},
\nl{Neon-Soaked Cyberpunk City},
\nl{Dragon’s Stormy Perch},
\nl{Reflections of Earth},
\nl{After The Fall},
\nl{Retro Gaming Nostalgia},
\nl{Medieval Village Life},
\nl{Samurai and the Mystical},
\nl{Minimalistic Geometry},
\nl{Alien Flora and Fauna},
\nl{The Inventor’s Steampunk Workshop}

\paragraph{ConceptNet concepts}
\nl{curling iron},
\nl{baseball stadium},
\nl{flowers},
\nl{submarine},
\nl{policeman},
\nl{projectile},
\nl{tissue holder},
\nl{jogging},
\nl{storey},
\nl{sickness},
\nl{parlor},
\nl{ships},
\nl{conductor},
\nl{booze},
\nl{key},
\nl{metal},
\nl{prostitute},
\nl{wings},
\nl{tools},
\nl{road},
\nl{main},
\nl{leader},
\nl{radio},
\nl{surprise},
\nl{chips},
\nl{castle},
\nl{bathroom},
\nl{compete against},
\nl{leather},
\nl{science},
\nl{rich},
\nl{sponge},
\nl{bell},
\nl{eloquent},
\nl{nightclub},
\nl{water},
\nl{patient},
\nl{eat vegetables},
\nl{respect},
\nl{lemur},
\nl{bum},
\nl{mammoth},
\nl{birthday},
\nl{chain},
\nl{cats},
\nl{frogs},
\nl{arkansas},
\nl{basketball},
\nl{listening},
\nl{dream},
\nl{ticket office},
\nl{failure},
\nl{text},
\nl{now},
\nl{oven},
\nl{leg},
\nl{mundane},
\nl{copulate},
\nl{tree},
\nl{wood},
\nl{mail},
\nl{wooden rod},
\nl{clippers},
\nl{competing against},
\nl{dull},
\nl{book},
\nl{watch television},
\nl{winning baseball game},
\nl{iphone},
\nl{dance club},
\nl{security},
\nl{politician},
\nl{subway station},
\nl{fall},
\nl{junk},
\nl{sleighing ride},
\nl{call},
\nl{mosquitoes},
\nl{national highway},
\nl{contraceptive device},
\nl{statement},
\nl{kill},
\nl{seeing old things},
\nl{lift},
\nl{adults},
\nl{pillowcase},
\nl{wedding ring},
\nl{eyes},
\nl{country},
\nl{stepladder},
\nl{mandolin},
\nl{reception area},
\nl{chief},
\nl{plastic},
\nl{projector},
\nl{hub},
\nl{card catalog},
\nl{negligible},
\nl{rook},
\nl{llano estacado}


\section{Robustness Experiments}
\label{app:robustness}

In the following sections, we conduct experiments to demonstrate our method's ability to provide robust concept interpretations. First, we show that the obtained decomposition is stable, \ie, that the same elements are learned across different choices of training sets and initialization. Second,  we test our decomposition's ability to generalize w.r.t. the denoising training task, \ie, we test the denoising capabilities of $w^*$ on test images. This shows that the elements in $w^*$ represent the entire concept, beyond the training images.

\subsection{Robustness to Training Data}
Note that since $N >> d$, there are many linear combinations that yield $w^*$. However, due to the specific MLP-based structure and the sparsity constraints, the decomposition is stable across multiple draws of training sets. In this section, we aim to verify that empirically.
For each concept, we generate $2$ alternative training sets with different random seeds, in addition to our original training set, to test the consistency of our results. For each alternative training set, we decompose the concept using our method as described in \cref{sec:method}, with a different initialization for the MLP\footnote{Our code does not employ a fixed random seed, thus each run implies a different random initialization for the MLP and a different set of random noises and timesteps for the training images.}. This process results in $3$ decompositions of $n=50$ tokens for each concept. 

We then analyze the intersection of the top $k=10, 25,$ and $50$ tokens between the original decomposition and each of the alternative decompositions.
The concept intersection score for $k$ is defined to be the average of the intersections with the two alternative sets. In other words, we calculate two intersection sizes for $k$: between the top $k$ tokens of the original decomposition and the first alternative decomposition, and between the top $k$ tokens of the original decomposition and the second alternative. The overall concept intersection score for $k$ is the average of the two. Standard Deviation is computed across the concepts. Note that this experiment measures an \emph{exact match} of the tokens, therefore the actual intersection may be even higher, \eg, if a synonym is used in one of the alternatives.

\begin{table}[h]

\centering
\setlength{\tabcolsep}{4pt}
\addtolength{\belowcaptionskip}{-0pt}
\caption{Decomposition consistency experiment. For each number of tokens ($k=10, 25, 50$) we test the intersection between our learned top tokens and those learned by employing two \emph{different} training sets of concept images, with different random initializations. The results demonstrate that the top tokens are consistent and robust across different training sets and seeds.}
\vspace{-0px}
    \begin{tabular}{cc@{\hspace{0.5cm}}c@{\hspace{0.5cm}}c}
    \toprule
        & No. of Tokens & Intersection \\
      \midrule
      \multirow{3}{*}{\begin{turn}{90}\small{Concrete}\end{turn}}&Top 10 & 8.03 (80.3$\%$)  $\pm$ 2.07 \\
      &Top 25 &  17.68 (70.7$\%$)  $\pm$ 4.47 \\
      &Top 50 &  28.96 (57.9$\%$) $\pm$ 8.05\\
      \midrule
      \multirow{3}{*}{\begin{turn}{90}\small{Abstract}\end{turn}}&Top 10 & 7.20 (72.0$\%$)  $\pm$ 1.86 \\
      &Top 25 &  15.95 (63.8$\%$)  $\pm$ 3.97 \\
      &Top 50 &  25.65 (51.3$\%$) $\pm$ 5.41\\
         \bottomrule
    \end{tabular}
    \label{tb:consistency}
    \vspace{-0px}
\end{table}

The average intersection scores across all concrete concepts and all abstract concepts are presented in \cref{tb:consistency}. As can be seen, for the concrete concepts, an average of $8.03 (80.3\%)$ of the top $10$ tokens are present in all the decompositions, even when considering an entirely different training set, indicating that the top tokens obtained by our method are stable. Additionally, when considering the top $25$ tokens, an average of $17.68 (70.7\%)$ of the tokens are present in all decompositions, which is a large majority. We note that the bottom tokens are typically less influential on the decomposition, as they are assigned relatively low coefficients by the MLP. Accordingly, when considering all $50$ tokens, an average of $28.96 (57.9\%)$ of the tokens appears in all decompositions. 
The results for the abstract concepts are slightly lower, yet demonstrate a similar behavior.
Overall, the results demonstrate that our method is relatively robust to different training sets and random seeds, such that even in the face of such changes, the top-ranked tokens remain in the decomposition.

\subsection{Denoising Generalization}
\begin{figure*}[h!]
    \centering
    \setlength{\tabcolsep}{0.5pt}
    \addtolength{\belowcaptionskip}{-0pt}
    {\small
    \begin{tabular}{c c}
        
        {\includegraphics[scale=0.35]{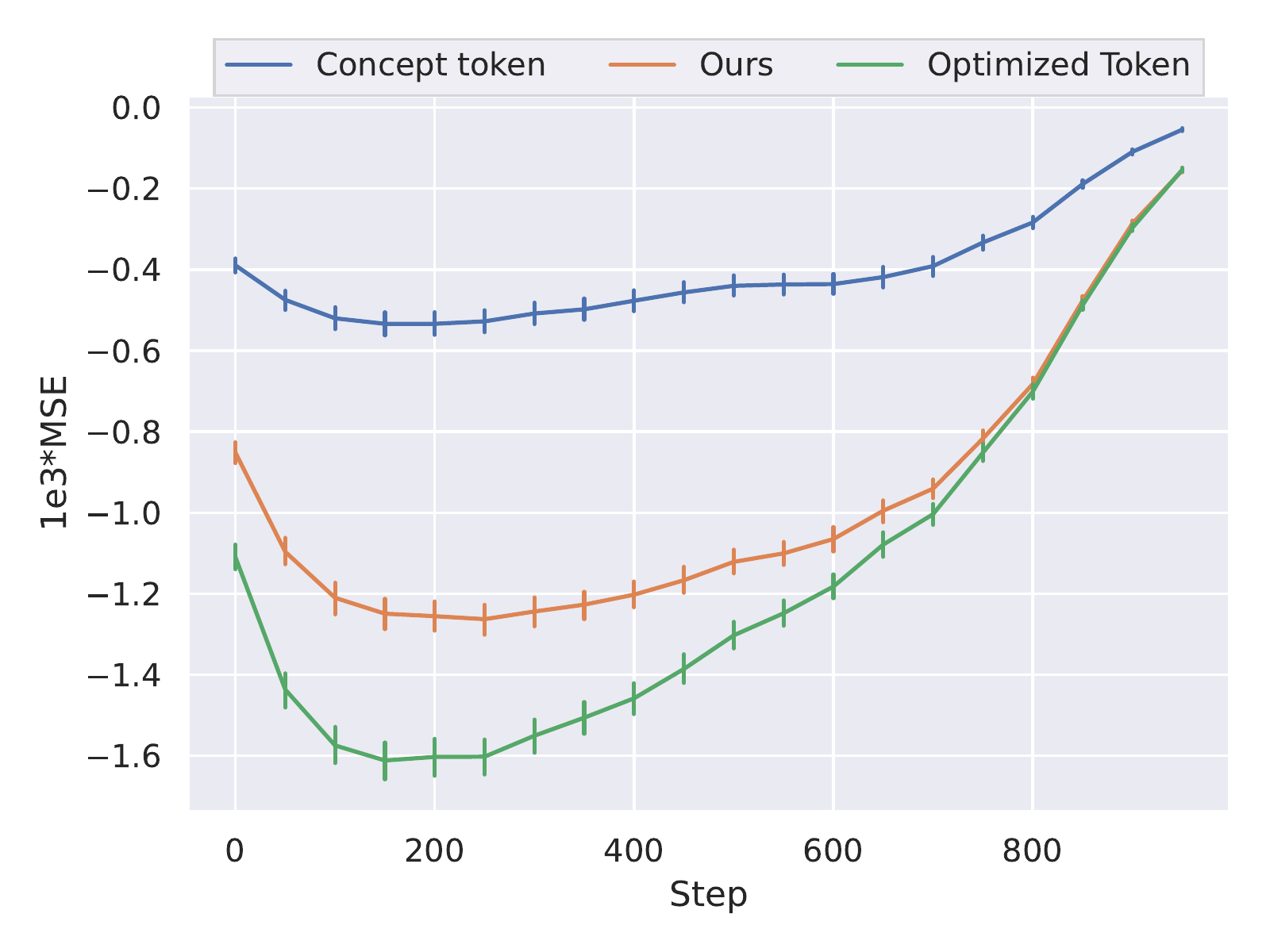}} \hspace{0.05cm} & \raisebox{+0.1\height}{\includegraphics[width=0.55\linewidth, clip]{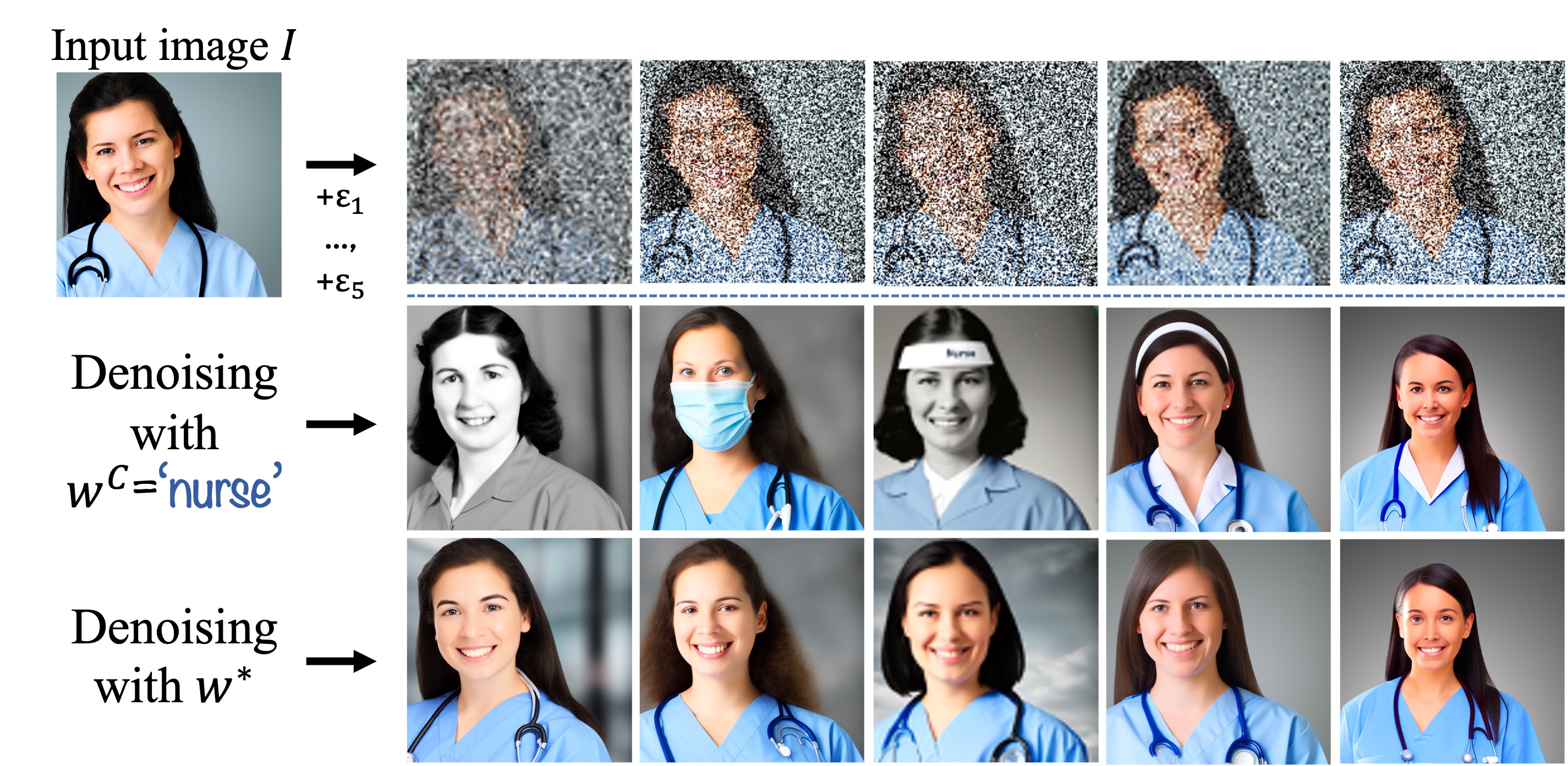}} \\[-0.2cm]
        (a) & (b)

    \end{tabular}
    
    }
    \vspace{-0px}
    \caption{Generalization tests comparing the concept prompt, $w^c$, and our pseudo-token, $w^*$. (a) Quantitative test on all concepts. For each timestep, we add random noises to the images and compare the reconstruction with ${w^*}$, $w^{c}$, and $w^o$, a continuous token optimized with the same objective as ${w^*}$ (Optimized Token). We report the MSE after subtracting the score of a random token.
    (b) Qualitative comparison. An image $I$ is generated from the concept \nl{a nurse}, and different noises are added $\varepsilon_1,\dots, \varepsilon_5$ (1st row). We then compare the denoising with $w^c$ (2nd row) and with $w^*$ (3rd row).}
    \label{fig:mse}
    \vspace{-0px}
\end{figure*}
\label{sec:exploration}
We wish to ascertain that the learned decomposition $w^*$ can generalize its training task to test images. Recall that during training, $w^*$ accumulated the elements necessary to denoise the concept images \emph{for each denoising step}. Next, we wish to demonstrate that these features can perform this same training task on test images.  We compare the denoising quality with $w^*$ to that of the concept prompt token(s), $w^{c}$. This comparison serves two purposes: (1) establish \textsc{Conceptor}'s ability to learn the actual concept features, beyond a set of features that simply reconstruct the training set, (2) motivate the fact that, as can be seen in our results, $w^* \neq w^c$.
To intuitively motivate the latter, note that even though $w^{c}$ was used to generate the concept images, it is not necessarily the best prompt to denoise them, 
since: (1) $w^c$ generates each image using a specific initial random noise, but is not guaranteed to be better in denoising them after applying other random noises. (2) Unlike $w^c$, $w^*$ is constructed as a \emph{linear combination of tokens}. Thus, our optimization is performed over a larger, \emph{continuous} space of embeddings, and therefore is more expressive than a simple selection of tokens.

We begin with a quantitative comparison. We sample a test set of $100$ images for each concept in the dataset. Then, for each denoising step $t\in \{1, \dots, T\}$ and each test image, we draw a random noise and apply it as in~Eq.~\ref{eq:latent}. Finally, we test the reconstruction loss specified in~Eq.~\ref{eq:reconstruction} with the pseudo-token ${w^*}$ compared to the concept prompt $w^{c}$. To provide a lower bound on the obtainable error, we additionally compare to $w^o$, a vector optimized with the same reconstruction objective on the entire continuous embedding space $\mathbb{R}^d$ without restrictions, similar to~\cite{gal2022image}. 
Note that, unlike ${w^*}$, $w^o$ does not offer interpretable information, since it is non-decomposable, and often out-of-distribution (see~\cref{sec:related}). However, it demonstrates our claim that optimization over a larger domain yields better reconstruction. Additionally, we observe that there is a large variance in the MSE score across timesteps. Latents in early steps are very noisy, and therefore obtain a very high loss ($\sim 0.8$), while the last steps contain virtually no noise, and the MSE is very low ($\sim 1e^{-3}$). Therefore, we compute a baseline score to normalize the scale. We subtract from each score the denoising score for the same images using a \emph{random token} which serves as an upper bound for the MSE.
\cref{fig:mse}(a) presents the results averaged across all concepts, showing that the concept $w^c$ obtains a score worse than both $w^*$ and the optimized token $w^o$, which obtains the best results. These differences are statistically significant, as shown by the error bars marked on every timestep. 
Evidently, by optimizing a token over a larger domain, we can outperform the original concept token in the denoising task. Importantly, this motivates the different features learned by $w^*$ while also demonstrating the remarkable denoising capability of $w^*$ on \emph{any concept image}, which indicates that $w^*$ indeed captures the concept features.

\cref{fig:mse}(b) provides a qualitative comparison between $w^c$ and $w^*$.
An input image $I$ generated by \nl{a photo of a nurse} is noised and then denoised back from different denoising steps, using the concept token $w^c$ and our pseudo-token $w^*$. 
As can be seen, there are cases where, given a different random seed, $w^c$ does not preserve the features in the original image $I$ (\eg, it adds hats, face masks, and black and white effects), while $w^*$ does. Intuitively, this can be attributed to the rich representation learned by $w^*$, which can include both semantic and style features. 
Both experiments motivate the diversity of the learned decomposition. Since $w^c$ is not necessarily optimal for Eq.~\ref{eq:reconstruction}, $w^*$ learns additional features to improve the denoising quality. 
Thus, $w^*$ balances two objectives-- interpretability and faithfulness to the model's internal representations.


\section{Single-image Decomposition}
\label{app:single_image}

\begin{figure*}[h!]
    \centering
    \addtolength{\belowcaptionskip}{0pt}
    \includegraphics[width=1\linewidth, clip]{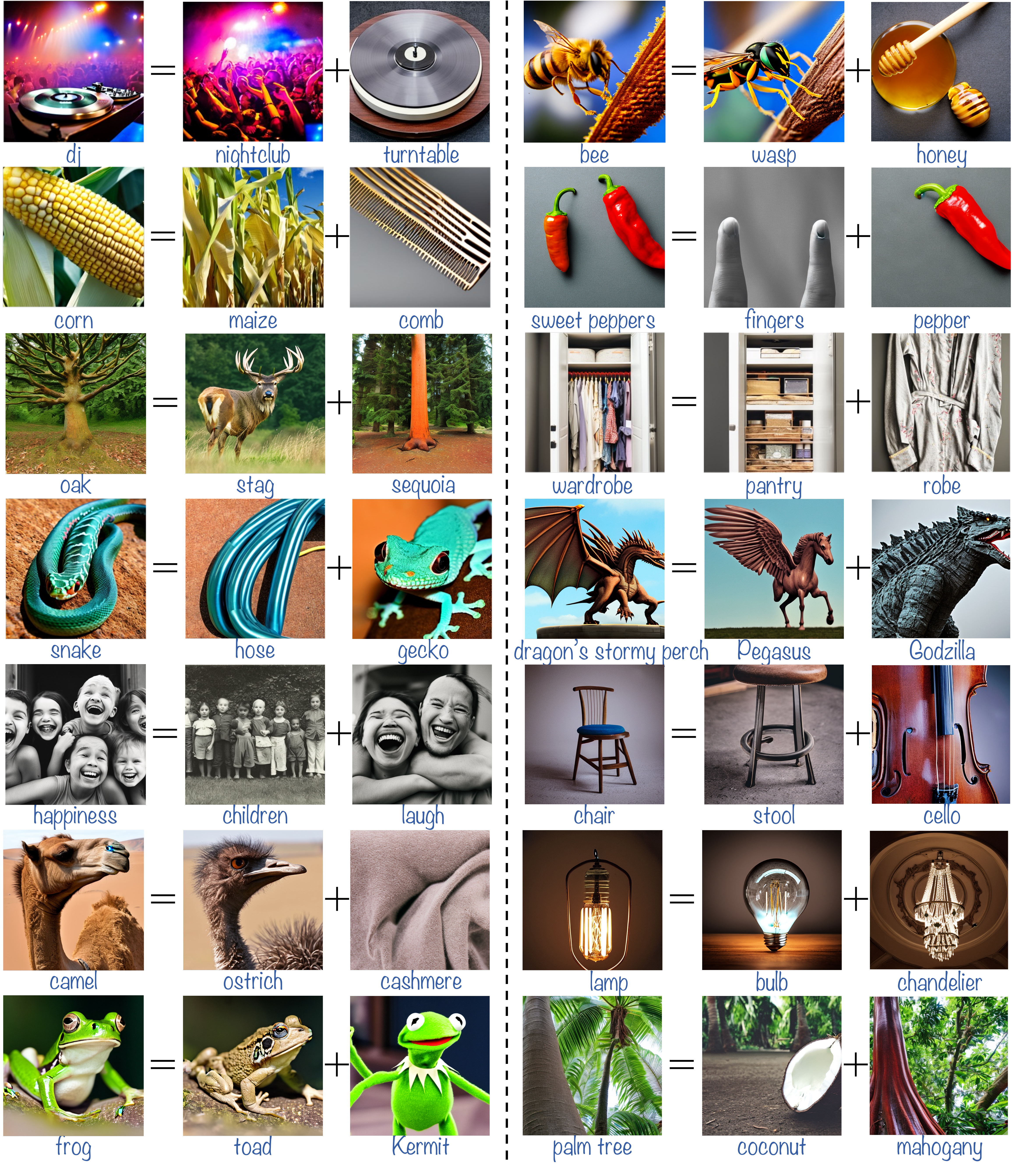}
    \caption{ 
    Decompositions of single images by \textsc{Conceptor}. Each of the examples depicts an image generated by Stable Diffusion for the concept, and its corresponding decomposition.
    }
    \vspace{-0px}
    \label{fig:single_image_supp}
\end{figure*}
In this section, we provide additional examples of single-image decompositions obtained by our method, as described in~\cref{sec:single_image} of the main paper. \cref{fig:single_image_supp} presents the obtained decompositions for images generated by Stable Diffusion for a given concept. As can be seen, the phenomena demonstrated in the main paper are reproduced in the presented decompositions. For example, \nl{sweet peppers} borrow the appearance from the \nl{pepper} and the shape from the \nl{fingers}, the \nl{frog} borrows the overall appearance of a \nl{toad} with the color of \nl{Kermit}, \etc. Other concepts are constructed as a composition of related elements, for example, \nl{happiness} is constructed as \nl{children} who are \nl{laughing}, and a \nl{lamp} is a \nl{chandelier} with a \nl{bulb} in it.

\subsection{Reliance on Renowned Artistic Styles}
\label{app:renown_styles}
\begin{figure}[h!]
    \centering
    \addtolength{\belowcaptionskip}{0pt}
    \includegraphics[width=0.7\linewidth, clip]{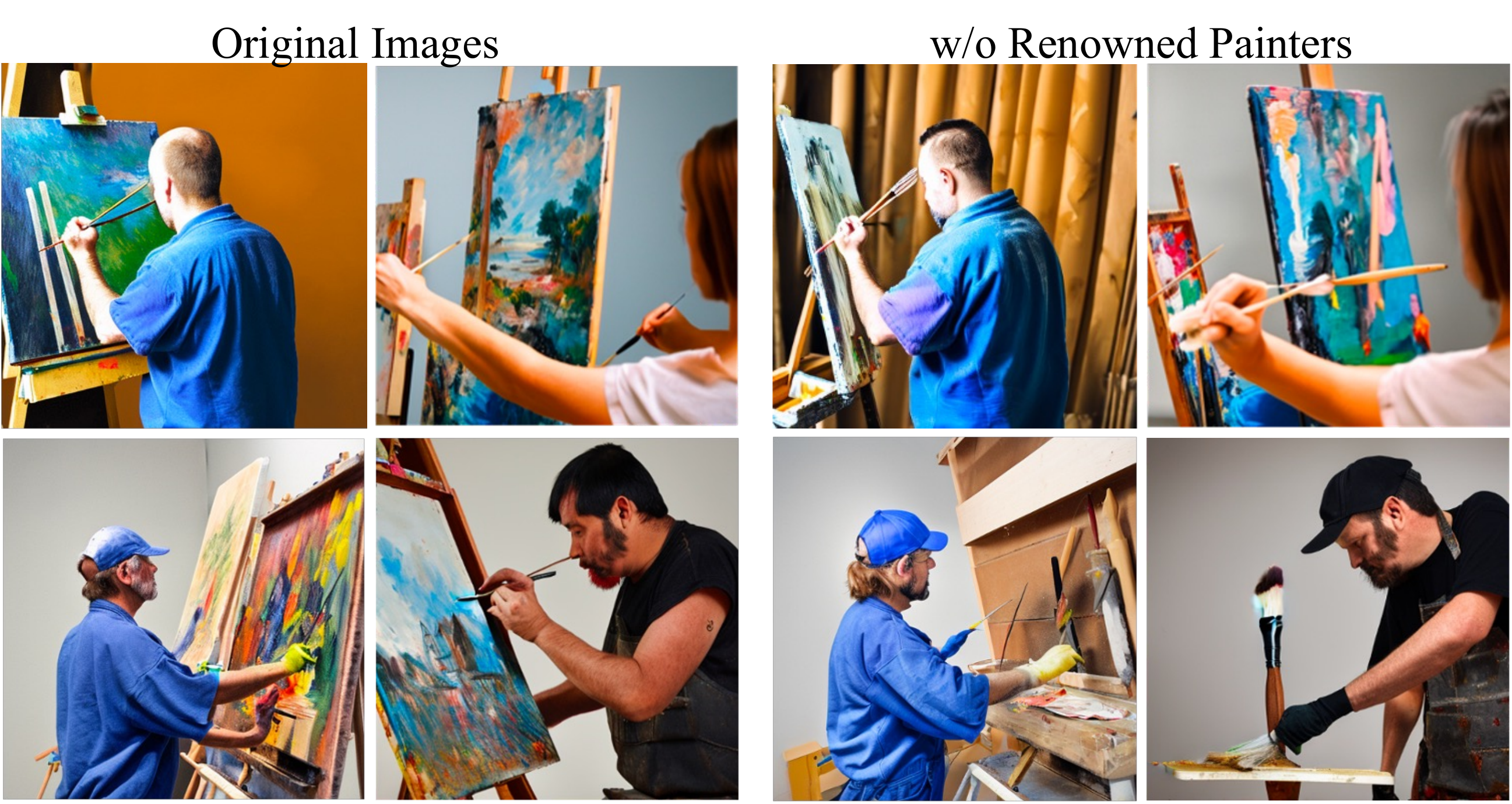}
    \caption{ 
    Generated images for the concept \nl{painter} before and after removing the names of famous painters from the single-image decomposition. As can be observed, this removal results in a significant modification to the painting in the image.
    }
    \vspace{-0px}
    \label{fig:renown_art}
\end{figure}

As demonstrated in~\cref{fig:teaser,fig:single_image}, we find that the representation of the concept \nl{painter} relies on the renowned artistic styles of famous artists. In this section, we further verify this observation through the lens of our single-image decomposition scheme. We generate $100$ test images for the concept \nl{painter} and apply our single-image decomposition scheme to all images. Out of the tested images, we found that $67$ images contain at least one name of a famous artist in their decomposition. This result empirically demonstrates the reliance on existing artistic styles, even \emph{when the prompt does not specify the artist's name explicitly}. \cref{fig:renown_art} demonstrates the impact of the experiment described above on the generated images. We observe that removing the names of renowned artists modifies the painting generated in the image entirely (top row) or, in some cases, removes it from the image altogether (bottom row).

\section{Representation by Exemplars}
\label{app:exemplars}
\begin{figure*}[t!]
    \centering
    \includegraphics[width=0.9\linewidth, clip]{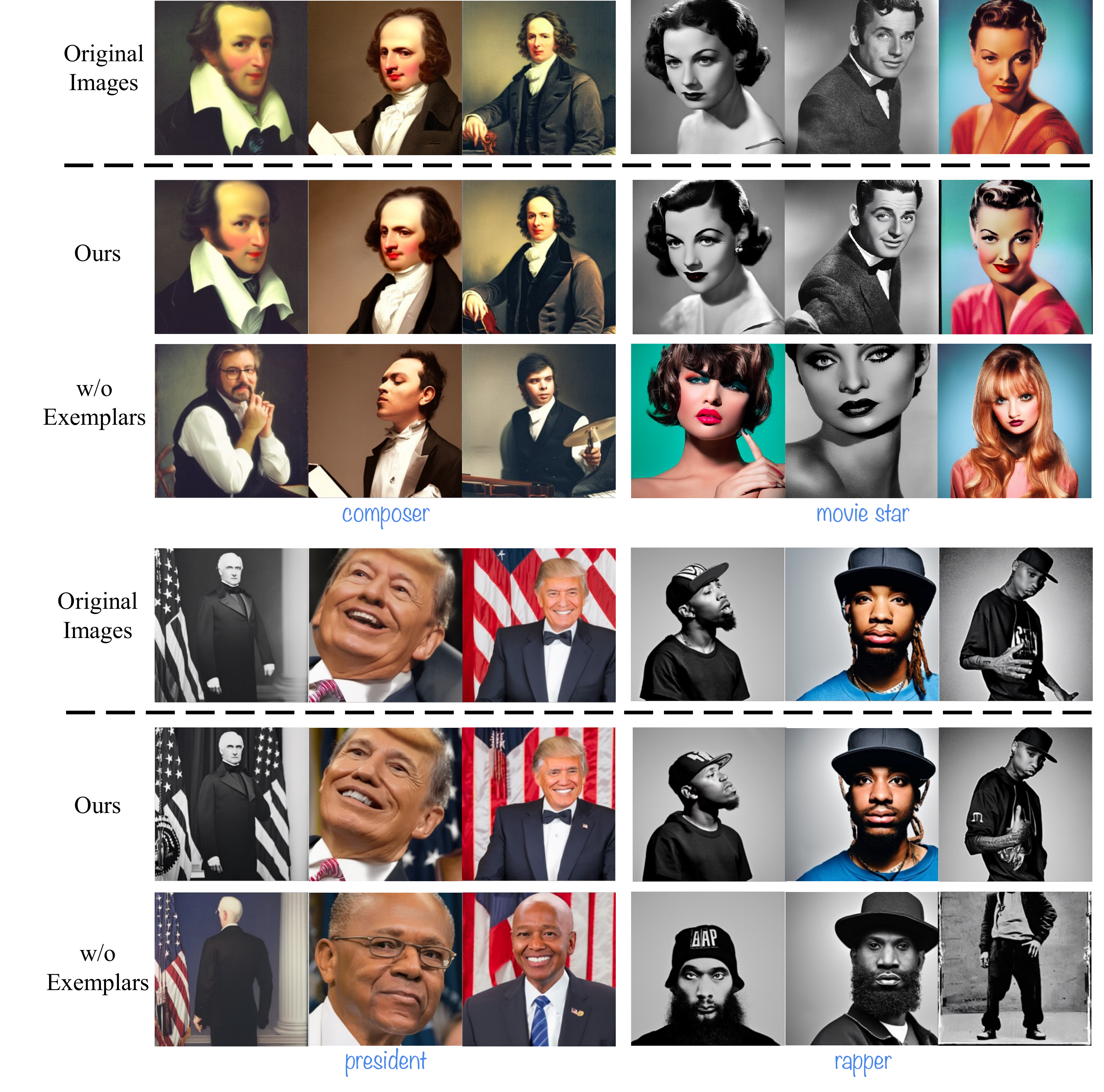}
    \vspace{-16px}
\caption{
     Examples of concepts that rely on famous instances. When removing the exemplars, the reconstruction quality is significantly harmed.
    }
    \vspace{-0px}
    \label{fig:exemplar}
\end{figure*}
\cref{tb:exemplars} and \cref{fig:exemplar} present examples of concepts that rely on famous instances for their representations. For example, \nl{movie star} is represented by names of famous actors such as Marilyn Monroe or Lucille Ball. Similarly, the concept \nl{president} is dominated by American presidents such as Obama and Biden, and the concept \nl{basketball player} relies on famous players such as Kobe Bryant and LeBron James.  To demonstrate the reliance on the exemplars, \cref{fig:exemplar} shows the reconstruction by our method with and without the famous instance names from~\cref{tb:exemplars}. As can be observed, the reconstruction quality heavily relies on the identities of the instances, and when we remove those instances the reconstruction is harmed significantly.
\begin{table*}[h!]
\small
\centering
\setlength{\tabcolsep}{4pt}
\addtolength{\belowcaptionskip}{-0pt}
\caption{Exemplar-based decomposition elements obtained by \textsc{Conceptor}.}
\vspace{-0px}
    \begin{tabular}{@{}l@{\hspace{0.8cm}} l @{\hspace{0.2cm}} l@{}}
    \toprule
      Concept & \textsc{Conceptor} \\
      \midrule
         Composer  &  \texttt{Schubert, Beethoven, Chopin, Mozart, Brahms, Wagner}\\
         Movie Star  &  \texttt{Aubrey, Bourne, Lucille, Gloria, Marilyn, Monroe, Oswald}\\
         President  &  \texttt{Obama, Trump, Biden, Nixon, Lincoln, Clinton, Washington}\\
         Rapper  &  \texttt{Tupac, Drake, Khalifa, Weekend, Khaled, Eminem, Wayne}\\
         Basketball player  &  \texttt{Kobe, LeBron, Shaq, Bryant, Jordan, Donovan, Kyrie}\\
         \bottomrule
    \end{tabular}
    \label{tb:exemplars}
    \vspace{-0px}
\end{table*}
\newpage
\clearpage

\section{Decomposition of Complex Concepts}
\label{app:complex_wordcloud}
\begin{figure*}[h!]
\begin{center}

    \includegraphics[width=0.8\linewidth, clip]{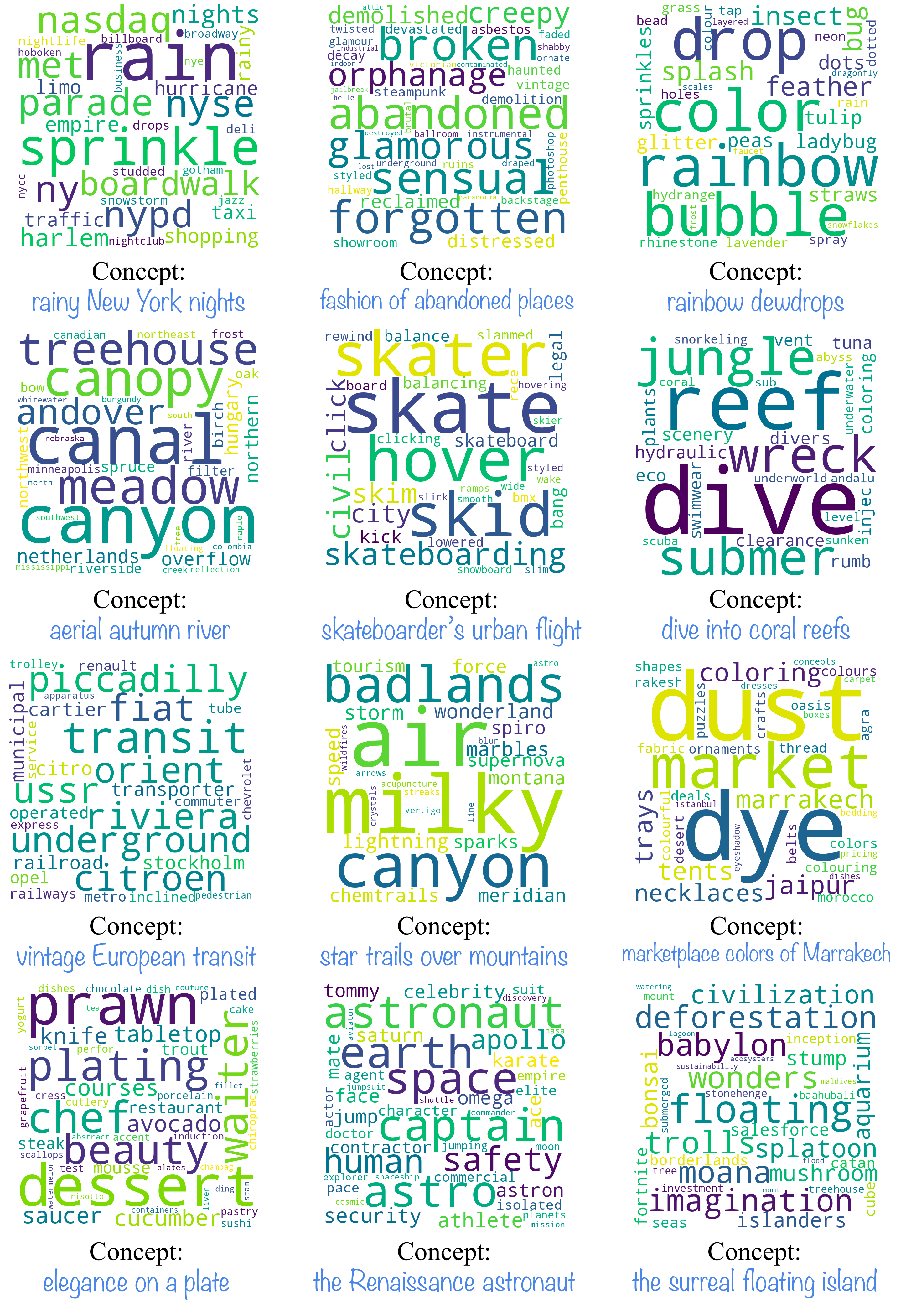}
    \end{center}
    \vspace{-16px}
    \caption{ 
     Decompositions obtained by \textsc{Conceptor} for the complex concepts from our dataset.
    }
    \label{fig:complex_wordclouds}
    \vspace{-0px}
\end{figure*}

In~\cref{fig:complex_wordclouds} we present the decomposition results for the first $12$ prompts from our complex prompts subset.
  
\newpage

\section{Concept Debiasing}
\label{app:debiasing}
\begin{figure*}[h!]
\begin{center}
    \includegraphics[width=0.85\linewidth, clip]{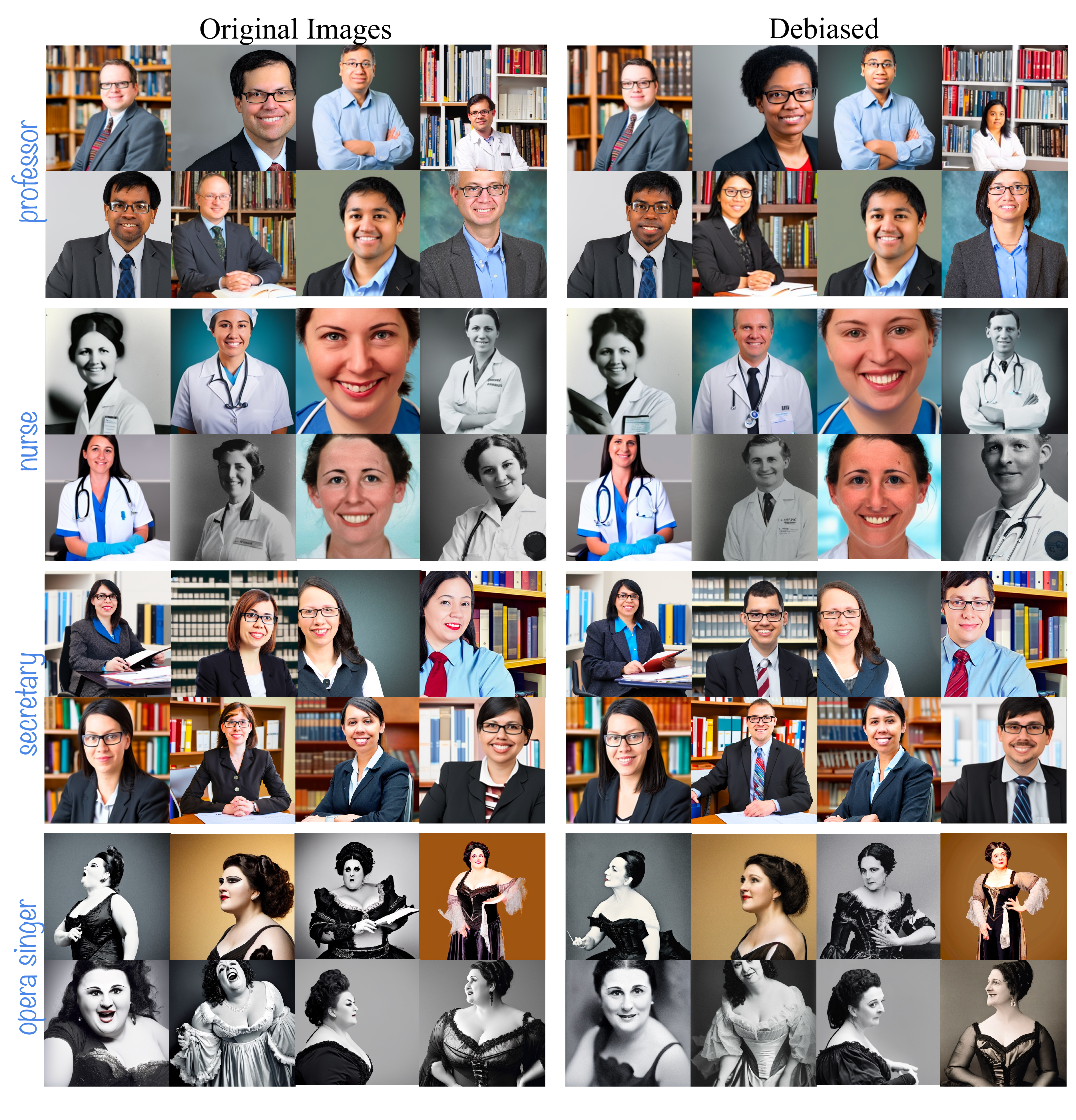}
\end{center}
    \vspace{-16px}
    \caption{ 
     Examples of concept debiasing using \textsc{Conceptor} with the same set of $8$ random seeds. The first three examples demonstrate gender debiasing, while the last one demonstrates a decoupling between \nl{opera singer} and \texttt{obesity}.
    }
    \label{fig:debias_supp}
    \vspace{-0px}
\end{figure*}

As mentioned in the main paper, \textsc{Conceptor} is capable of detecting biases that are otherwise difficult to capture by simply observing the generated images. After detecting a bias, one could employ a coefficient manipulation with our method, or opt to employ other methods for concept editing such as~\cite{gandikota2023unified}, based on the information provided by \textsc{Conceptor}.
\cref{fig:debias_supp} presents examples of concept debiasing by employing coefficient manipulation with \textsc{Conceptor}. For each concept (row), we decrease the coefficient of the biased tokens in the decomposition until an unbiased representation is achieved. \cref{fig:debias_supp} demonstrates a comparison of the generated images by SD without intervention (Original Images), and the images reconstructed by our method after reducing the bias coefficients, on the same set of $8$ random seeds. As can be observed, our method is able to mitigate the different biases while maintaining the other features represented by the concept.

\newpage

\section{Principal Component Visualization for Baselines}
\label{app:componenet}
\begin{figure*}[h!]
\begin{center}
    \includegraphics[width=0.87\linewidth, clip]{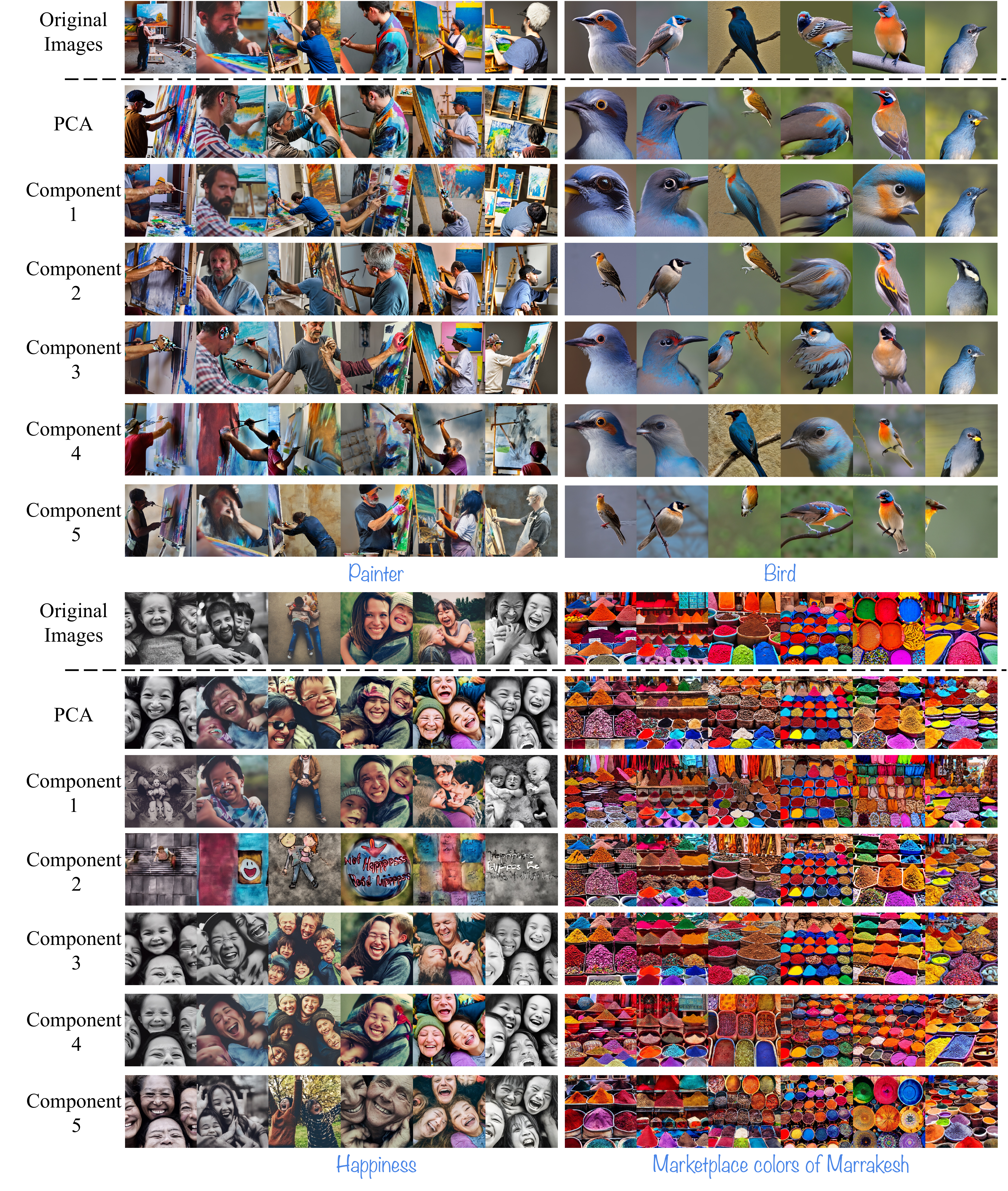}
\end{center}
    \vspace{-16px}
    \caption{ 
     PCA top $5$ extracted principal components for $4$ concepts. The first row depicts SD's original images, the second row shows the reconstruction by PCA, and the last $5$ rows demonstrate the impact of removing each of the top $5$ principal components learned by PCA.
    }
    \label{fig:pca_components}
    \vspace{-12px}
\end{figure*}

In this section, we enclose the concept activation vectors (CAVs) learned by the leading concept-based explainability method adapted from CNNs (PCA). \cref{fig:pca_components} presents the reconstruction and components learned by PCA. To visualize a component, we remove it and generate the corresponding images without it to examine its impact on the generations. As can be seen, most of the learned components are not easily interpretable and do not demonstrate a coherent semantic change across the different test images. Thus, these methods violate the \emph{meaningfulness} criterion. We note that meaningfulness is a critical criterion for an interpretation, as without it, humans cannot understand the decomposition, and it does not serve its basic goal of giving insights into the model.


\end{document}